\documentclass{article}

 \usepackage[preprint]{neurips_2026} % "preprint" option is used for arXiv or other preprint submissions !!!!!!

\usepackage[utf8]{inputenc} % allow utf-8 input
\usepackage[T1]{fontenc}    % use 8-bit T1 fonts
\usepackage{hyperref}       % hyperlinks
\usepackage{url}            % simple URL typesetting
\usepackage{booktabs}       % professional-quality tables
\usepackage{amsfonts}       % blackboard math symbols
\usepackage{nicefrac}       % compact symbols for 1/2, etc.
\usepackage{microtype}      % microtypography
\usepackage{xcolor}         % colors
\usepackage{graphicx}
\usepackage{subcaption}
\usepackage{float}
\usepackage{amsmath}
\usepackage{comment}
\usepackage{multirow}
\usepackage[most]{tcolorbox}
\usepackage{pifont} %for the \ding symbols and 
\usepackage[table]{xcolor}
\newcommand{\yes}{\textcolor{green!60!black}{\ding{51}}}
\newcommand{\no}{\textcolor{red!70!black}{\ding{55}}}

\usepackage{enumitem}
\setlist{nosep}

\usepackage[font=small,labelfont=bf,skip=4pt]{caption}
\usepackage{subcaption}

\title{\raisebox{-0.3\height}{\includegraphics[height=1.3em]{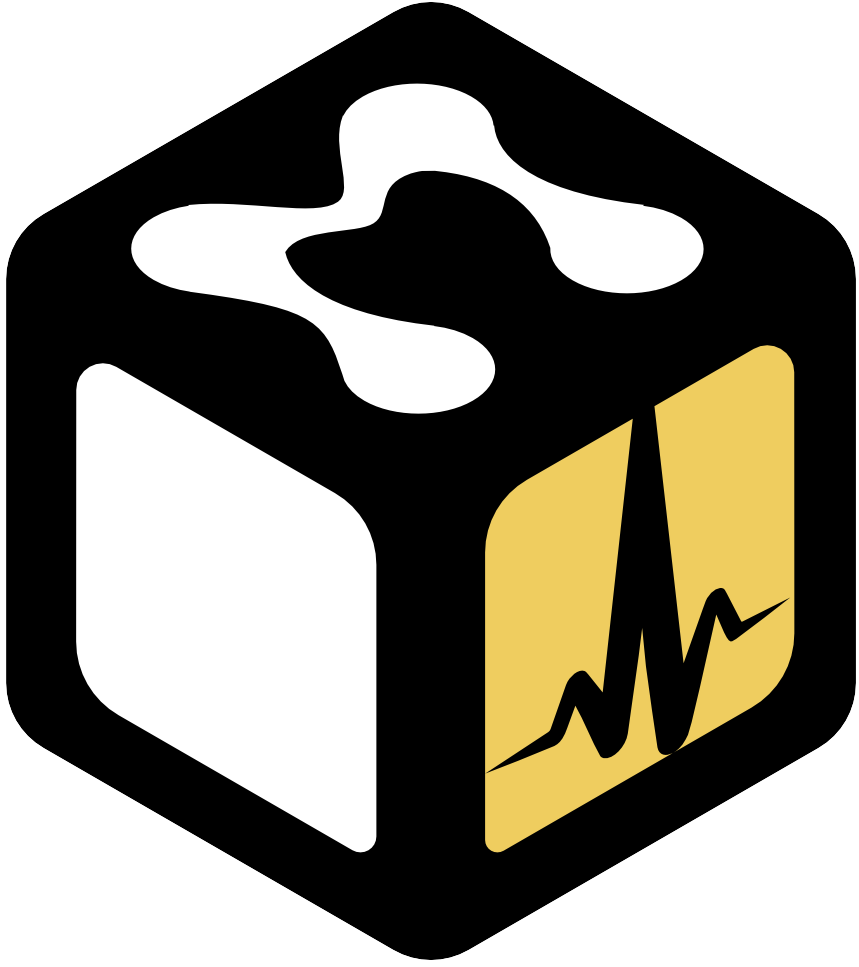}}\,Fully Open Meditron: An Auditable Pipeline for Clinical LLMs}
\author{%
  Xavier Theimer-Lienhard\thanks{Corresponding author: \texttt{xavier.theimer-lienhard@epfl.ch} $^{\dagger}$ Equal supervision.} \And
  Mushtaha El-Amin \AND
  Fay Elhassan \And
  Sahaj Vaidya \And
  Victor Cartier-Negadi \AND
  David Sasu$^{\dagger}$ \And
  Lars Klein$^{\dagger}$ \And
  Mary-Anne Hartley$^{\dagger}$ \And
  \\
  LiGHT, EPFL \\
  \textbf{Models:} \url{https://huggingface.co/collections/EPFLiGHT/meditronfo} \\
  \textbf{Dataset:} \url{https://huggingface.co/datasets/EPFLiGHT/fully-open-meditron} \\
  \textbf{Code:} \url{https://github.com/EPFLiGHT/FullyOpenMeditron} \\
}

\begin{document}

\maketitle

\begin{abstract}
Clinical decision support systems (CDSS) require scrutable, auditable pipelines that enable rigorous, reproducible validation. Yet current LLM-based CDSS (LLM-CDSS) remain largely opaque. Most ``open'' models are open-weight only, releasing parameters, while withholding the data provenance, curation procedures, and generation pipelines that determine model behavior. Fully Open (FO) models, which expose the complete training stack end-to-end, do not currently exist in medicine. 
We introduce Fully Open Meditron, the first fully open pipeline for building LLM-CDSS, comprising a clinician-audited training corpus, a reproducible data construction and training framework, and a use-aligned evaluation protocol. The corpus unifies eight public medical QA datasets into a normalized conversational format and expands coverage with three clinician-vetted synthetic extensions: exam-style QA, guideline-grounded QA derived from 46,469 clinical practice guidelines, and clinical vignettes. 
The pipeline enforces system-wide decontamination to eliminate overlap with evaluation benchmarks, includes gold-label resampling of teacher generations, and also includes end-to-end validation by a four-physician panel. 
We evaluate using an LLM-as-a-judge protocol over expert-written clinical vignettes, calibrated against 204 human raters, capturing open-ended clinical reasoning beyond typical multiple-choice benchmarks. We apply the recipe to five FO base models (Apertus-70B/8B-Instruct, OLMo-2-32B-SFT, EuroLLM-22B/9B-Instruct). All MeditronFO variants are preferred over their bases in pair-wise clinical evaluation. Apertus-70B-MeditronFO improves +6.6 points over its base (47.2\% → 53.8\%) on aggregate medical benchmarks, establishing a new FO SoTA for LLM-CDSS. Additionally, Gemma-3-27B-MeditronFO is preferred over MedGemma in 58.6\% of LLM-as-a-judge comparisons and also outperforms it on HealthBench (58\% vs 55.9\%). These results show that fully open pipelines can achieve state-of-the-art domain-specific performance, without sacrificing auditability or reproducibility. 

\end{abstract}

\begin{figure}[h]
    \centering
    \includegraphics[width=\textwidth]{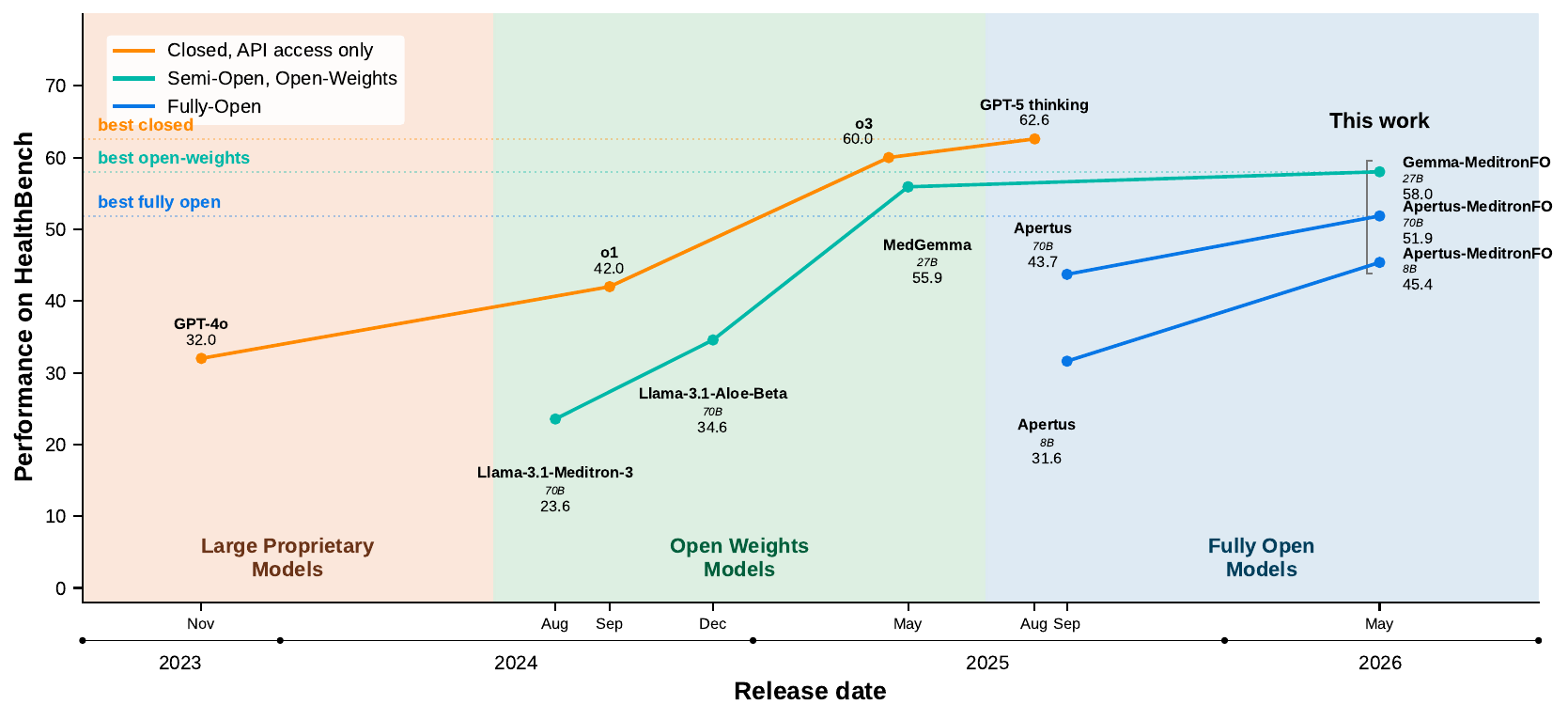}
    \caption{\textbf{Evolution of medical LLM performance on Healthbench over time across closed-data, open-weight, and fully open models.} While open-weight medical specialists have approached the performance of proprietary systems, no fully open medical specialist previously existed. This work introduces Apertus-MeditronFO, the first fully open medical specialist model, establishing a new state of the art among fully open systems.}
    \label{fig:main}
\end{figure}

\section{Introduction}
Medical large language models (LLMs) are increasingly being deployed in high-stakes clinical settings, from specialist decision support to autonomous patient-facing applications that may operate with little external scrutiny. As these systems encounter increasingly variable real-world interactions, questions of trust, auditability and provenance become increasingly important. Yet most ``open'' medical LLMs release only model weights while withholding the training data provenance, data preparation pipelines, and adaptation procedures that shape model behavior.  
Adapting generalist large language models into medical specialists is now widespread, producing systems such as MedGemma \citep{sellergren2025medgemma}, Aloe Beta \citep{garcia2026aloe}, Meditron \citep{chen2023meditron}, and BioMistral \citep{labrak2024biomistral}. The typical pipeline combines continued pre-training on medical corpora with supervised fine-tuning on curated QA datasets. However, the resulting systems remain largely opaque. Releasing weights alone does not reveal whether a model learned from guideline-grounded evidence, benchmark artifacts, synthetic hallucinations, or clinically narrow populations. Consistent with concerns raised about opaque adaptation pipelines \citep{alber2025medical}\citep{betley2026training}, current open-weight specialists including MedGemma do not disclose training corpora or generation pipelines, limiting independent auditability.

This concern is amplified by the saturation of standard medical benchmarks, where performance gains may reflect contamination, memorization, or benchmark-specific adaptation rather than clinical capability. In clinical practice, where clinicians, regulators, and patients may reasonably demand to audit what a model has learned
and how it was trained, this opacity presents a fundamental limitation. Fully open (FO) models offer a path to end-to-end auditability, but also operate under a disadvantage: because training data, preparation pipelines, and model weights must be openly releasable, they cannot rely on proprietary clinical corpora, restricted datasets, or undisclosed synthetic pipelines that underpin many frontier systems. As a result, FO models generally lag behind closed-data counterparts on established benchmarks, and no fully open medical specialist currently exists.

\begin{table*}[h]
\centering
\caption{\textbf{Openness dimensions across medical LLMs.} Most prior medical LLMs release weights but withhold the data and pipelines that determine model behavior. MeditronFO is the first family to satisfy all openness dimensions end-to-end. Openness is assessed separately for the \textit{base model} and the \textit{medical adaptation}. For the base model, \textbf{Data} refers to pretraining, post-training, instruction-tuning, or alignment data; \textbf{Code} refers to reproducible training code and recipe; and \textbf{Weights} refers to released model weights. For the medical adaptation, \textbf{Data} refers to fine-tuning, instruction and synthetic data; \textbf{Synth} refers to the synthetic data generation pipeline code, including prompts, teacher model, and filtering procedure; \textbf{Code} refers to the fine-tuning/training code and recipe and \textbf{Weights} refers to the adapted medical model weights. License categories are \textbf{O}=permissive open license, \textbf{C}=community or commercially usable license with restrictions, \textbf{IC}=inherited C license, IC reflects the base model's license; the medical adaptation itself is permissively released, and \textbf{R}=restrictive, research-only, or proprietary license.}
\label{tab:fully_open_comparison}
\resizebox{\textwidth}{!}{%
\begin{tabular}{llccc cccc c}
\toprule
\textbf{Model} & \textbf{Base} & \multicolumn{3}{c}{\textbf{Base model openness}} & \multicolumn{4}{c}{\textbf{Medical adaptation openness}} & \textbf{License} \\
\cmidrule(lr){3-5} \cmidrule(lr){6-9}
& & \textbf{Data} & \textbf{Code} & \textbf{Weights} & \textbf{Data} & \textbf{Synth} & \textbf{Code} & \textbf{Weights} & \\
\midrule
\multicolumn{10}{l}{\textit{Closed}} \\
\addlinespace[2pt]
Med-Gemini & Gemini & \no & \no & \no & \no & \no & \no & \no & R \\
\addlinespace[2pt]
\midrule
\addlinespace[2pt]
\multicolumn{10}{l}{\textit{Open Weights}} \\
\addlinespace[2pt]
BioMistral & Mistral & \no & \no & \yes & \yes & \no & \no & \yes & O \\
Meditron-70B & Llama 2 & \no & \no & \yes & \no & \no & \yes & \yes & IC \\
Meditron-3 & Llama 3.1 & \no & \no & \yes & \no & \no & \no & \yes & IC \\
Aloe Beta & Llama 3.1 & \no & \no & \yes & \yes & \no & \no & \yes & R \\
MedGemma-27B & Gemma-3-27B-it & \no & \no & \yes & \no & \no & \no & \yes & C \\
\addlinespace[2pt]
\midrule
\addlinespace[2pt]
\multicolumn{10}{l}{\textit{Partially Open}} \\
\addlinespace[2pt]
\rowcolor{gray!15} \textbf{Gemma-3-27B-MeditronFO} & Gemma-3-27B-it & \no & \no & \yes & \yes & \yes & \yes & \yes & IC \\
\addlinespace[2pt]
\midrule
\addlinespace[2pt]
\multicolumn{10}{l}{\textit{Fully Open}} \\
\addlinespace[2pt]
\rowcolor{gray!15} \textbf{Apertus-70B-MeditronFO} & Apertus-70B & \yes & \yes & \yes & \yes & \yes & \yes & \yes & O \\
\rowcolor{gray!15} \textbf{OLMo-2-32B-MeditronFO} & OLMo-2-32B & \yes & \yes & \yes & \yes & \yes & \yes & \yes & O \\
\rowcolor{gray!15} \textbf{EuroLLM-22B-MeditronFO} & EuroLLM-22B & \yes & \yes & \yes & \yes & \yes & \yes & \yes & O \\
\bottomrule
\end{tabular}}
\end{table*}

We argue that this gap reflects corpus construction rather than an inherent limitation of open models. Public medical benchmarks are heterogeneous, narrowly scoped, and poorly aligned with clinical interaction; for instance, emergency-care scenarios account for only 15\% of the aggregated public QA we examine, and life-threatening cases for under 9\%, despite being the settings where clinical decision support matters most. Prior work shows that biomedical specialists frequently fail to outperform their generalist bases on unseen medical data, suggesting reported gains may reflect contamination or benchmark adaptation rather than clinical capability ~\citep{dorfner2024biomedical}. Existing benchmarks also underrepresent low-resource settings, vulnerable populations, and diagnostic reasoning under uncertainty.
This issue is further exacerbated by the over-reliance on multiple-choice evaluation. MCQs reward rote structural recall but fail to capture clinically important dimensions, such as contextual awareness, communication, harmlessness, and alignment with guidelines. Building clinically useful models therefore requires open-ended evaluation and training corpora designed accordingly.

\textbf{Contributions.} To address this gap, we introduce \textbf{Fully Open Meditron}, the first FO pipeline for adapting FO foundation models into medical specialists. We show that competitive medical specialization can be achieved under FO constraints through disciplined clinician-audited corpus construction and open-ended clinical evaluation. Our main contributions are:

\noindent \textbullet~ \textbf{A fully open medical adaptation framework.} We release a reproducible end-to-end framework spanning corpus construction, synthetic data generation, decontamination, training, and evaluation for adapting fully open foundation models to medicine.

\noindent \textbullet~ \textbf{A structured, fully open clinician-audited knowledge corpus.} We normalize eight public medical QA datasets and systematically expand coverage via clinician-vetted synthetic generation, shifting emergency-care coverage from 15.0\% to 38.7\% and life-threatening severity from 8.6\% to 31.8\% (exam-style QA, guideline-grounded QA derived from 46,469 clinical practice guidelines, and open-ended clinical vignettes seeded from a unique global scale clinical evaluation corpus). The pipeline enforces rigorous decontamination and utilizes gold-label resampling of synthetic targets.

\noindent \textbullet~ \textbf{An automated, open-ended clinical evaluation protocol.} We introduce Auto-MOOVE, an LLM-as-a-judge framework validated against 204 human raters to assess multidimensional clinical reasoning beyond standard MCQ metrics.

\noindent \textbullet~ \textbf{A family of fully open medical specialists.} We apply this recipe to five fully open base models spanning three model families. Apertus-70B-MeditronFO improves from 47.2\% to 53.8\% on aggregate medical benchmarks, establishing a new fully-open SoTA. In open-ended evaluations, Gemma-3-27B-MeditronFO is preferred over MedGemma on Auto-MOOVE and scores higher on HealthBench, suggesting that the pipeline improves dimensions not captured by MCQA alone.

\section{Related works}
\label{sec:related}

\textbf{Open and fully open medical LLMs.}
Closed-data specialists including the MedPaLM family~\citep{singhal2023large, singhal2025toward} and Med-Gemini~\citep{saab2024capabilities} report strong medical benchmark performance but disclose neither training corpora nor adaptation pipelines. In parallel, a growing body of work adapts open-weight generalist LLMs into medical specialists. HuatuoGPT-II~\citep{chen2023huatuogpt} unifies pretraining and fine-tuning into a single stage, while MC-LLaMA~\citep{wu2024pmc} and BioMistral~\citep{labrak2024biomistral} continue-pretraining on biomedical corpora before instruction-tuning on aggregated QA benchmarks.  Meditron-70B~\citep{chen2023meditron,sallinen2025llama} scales this recipe with curated clinical guidelines. Despite growing interest in openness, most medical LLMs remain only partially transparent: often releasing weights, subsets of training data, or benchmark recipes, while withholding key components such as data provenance, filtering procedures, synthetic generation pipelines, or adaptation workflows. Even open-weight systems such as MedGemma~\citep{sellergren2025medgemma} disclose neither their training data nor their synthetic-generation pipelines. A detailed comparison of openness dimensions across all models is provided in
Appendix~\ref{appendix:fully_open_comparison}.

\textbf{Risks of opaque adaptation pipelines.}
Recent work demonstrates that medical LLMs are vulnerable both to targeted corruption of adaptation data and to broader behavioral drift induced by narrow-domain fine-tuning. ~\citep{alber2025medical} showed fine-tuning attacks that survive standard safety evaluations, while Betley et al.~\citep{betley2026training} show that fine-tuning on narrow corruptions in one domain can induce broadly misaligned deployment behavior.

\textbf{Benchmark contamination and decontamination.}
\citep{deng2024investigating} demonstrate substantial overlap between widely used evaluation benchmarks (MMLU ~\citep{hendrycks2020measuring}, TruthfulQA ~\citep{lin2022truthfulqa}, HellaSwag ~\citep{zellers2019hellaswag}, WinoGrande~\citep{sakaguchi2021winogrande}, GSM8K~\citep{cobbe2021training}, OpenBookQA~\citep{mihaylov2018can}) and major pretraining corpora, using both retrieval-based search and a Testset Slot Guessing protocol applicable to open- and closed-weight models. \citep{golchin2023time} complement this by showing that contamination can be detected post-hoc through prompting strategies that elicit verbatim recall of evaluation instances. Fully Open Meditron mitigates this risk through the two-stage n-gram and token-alignment decontamination pipeline introduced in Apertus~\citep{apertus2025}, applied system-wide against all evaluation references regardless of training-component provenance.

\textbf{Clinician participation in the development of open medical AI.}
Med-PaLM~\citep{singhal2023large} introduced multi-axis physician evaluation across factuality, reasoning, possible harm, and bias, and HealthBench~\citep{arora2025healthbench} scaled this to 5{,}000 physician-authored conversational rubrics. Thirunavukarasu et al.~\citep{thirunavukarasu2023large} similarly argue that clinical deployment requires evaluation paradigms grounded in workflows rather than exam-style recall. Fully Open Meditron incorporates clinician input at both the data-curation and evaluation stages, with a four-physician panel auditing synthetic-generation prompts and Auto-MOOVE built on expert-written vignettes.

\textbf{Open-ended evaluation at scale.} Recent work addresses the limitations of multiple-choice evaluation through rubric-based protocols: HealthBench~\citep{arora2025healthbench} scores model responses against physician-authored rubrics across thousands of conversational scenarios, and LiveClin~\citep{wang2026liveclin} introduces an updated benchmark to mitigate contamination. Pairwise preference evaluation has emerged as a complementary paradigm, both in domain-specific settings such as MOOVE~\citep{sallinen2025llama}, which collects expert comparisons over clinical vignettes, and in platforms such as Chatbot Arena~\citep{zheng2023judging}, which aggregates large-scale human pairwise judgments into model rankings. These approaches highlight that relative comparisons are often more reliable than absolute scoring, but rely heavily on human annotation, limiting scalability in specialised domains. LLM judges offer a path to scalable pairwise evaluation: \citep{zheng2023judging} establishes the paradigm and shows GPT-4 matches expert crowd preferences on open-ended dialogue, while \citep{thakur2024judging} shows that Cohen's $\kappa$ is a more reliable validation metric than raw percent agreement, and Han et al.~\citep{han2025judge} introduce a human-likeness test that situates a judge's $\kappa$ within the distribution of per-rater $\kappa$ values from a human panel. The MOOVE platform~\citep{sallinen2025llama} collects pairwise expert preferences over clinical vignettes; Auto-MOOVE builds on this by automating the comparison protocol with an LLM-as-a-judge validated against human inter-rater agreement.

%%%%%%%%%%%%%%%%%%%%%%%%%%%%%%%%%%%%%%%%%%%%%%%%%%%%%%%%%%%%
% THE Fully Open Meditron CORPUS
%%%%%%%%%%%%%%%%%%%%%%%%%%%%%%%%%%%%%%%%%%%%%%%%%%%%%%%%%%%%

\section{The Fully Open Meditron Corpus}
\vspace{-0.2cm}
\begin{figure}[h]
    \centering
    \includegraphics[width=\textwidth]{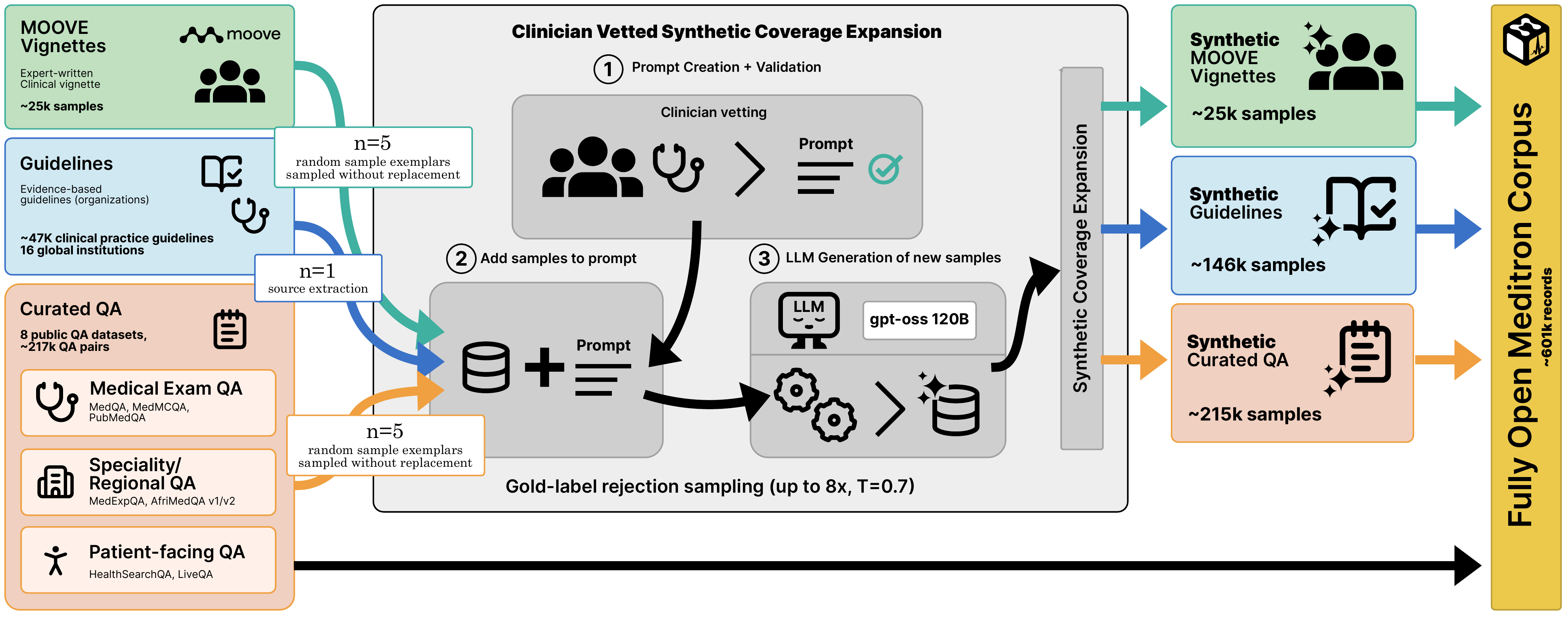}
    \vspace{-0.2cm}
    \caption{\textbf{The Fully Open Meditron Corpus construction pipeline.} The corpus combines three source streams: (1) eight aggregated public medical QA datasets (Curated QA), (2) 46,469 clinical practice guidelines from 16 global institutions (GUIDELINES), and (3) Expert-written clinical vignettes from the MOOVE training split. Clinician-vetted prompts and sampled exemplars are passed to GPT-OSS-120B to generate three synthetic components: Synthetic Curated QA (novel exam-style QA pairs, stratified by question type), Synthetic Guidelines QA (guideline-grounded QA), and Synthetic MOOVE (novel open-ended clinical vignette prompts designed to elicit complex diagnostic reasoning). Hallucinations are mitigated via gold-label rejection-sampling. Source and synthetic components are merged into the final Fully Open Meditron Corpus.}
    \label{fig:second}
\end{figure}

\subsection{Data Aggregation}
The foundation of our Fully Open Meditron Corpus is an aggregation of eight public medical QA datasets. To capture both exam-style reasoning and open-ended clinical interaction, we unify MedQA ~\citep{jin2021disease}, MedMCQA ~\citep{pal2022medmcqa}, PubMedQA ~\citep{jin2019pubmedqa}, MedExpQA ~\citep{alonso2024medexpqa}, HealthSearchQA ~\citep{singhal2023large}, and LiveQA ~\citep{abacha2017overview}. We additionally include AfriMed-QA v1/v2 ~\citep{olatunji2024afrimed} to partially mitigate the North American and European bias of standard medical benchmarks and expand representation of diverse clinical settings. Only training splits are utilized. All entries are normalized into a unified $system, user, assistant$ conversational format incorporating step-by-step rationales, discarding items that cannot be unambiguously mapped. This harmonization preserves diagnostic reasoning trajectories across heterogeneous source formats spanning MCQA, consumer-health queries, and open-ended specialist examinations and aligns with principles described in the MedGemma technical report ~\citep{sellergren2025medgemma}; dataset sources and sizes summarized in Table \ref{tab:training_datasets}.
Because medical benchmarks are widely contaminated in pretraining corpora, we apply a system-wide decontamination against all evaluation references, adapting the two-stage n-gram and token-alignment pipeline from Apertus. The reference set spans all benchmarks used in this work: MedQA, MedMCQA, PubMedQA, MedXpertQA, MMLU-Pro, IFEval, and ARC-Challenge. Specific thresholds, implementation details in Appendix~\ref{appendix:decontamination}.

To characterize this curated corpus and identify coverage gaps, we use an LLM as a zero-shot clinical metadata extractor over the first turn of each conversation. Extracted attributes include geographic context, resource setting, level of care, clinical severity, medical specialty, question type, and patient demographics. This analysis reveals that naïvely aggregating public benchmarks underrepresents clinically important dimensions such as low-resource settings, pediatric and geriatric populations, and open-ended diagnostic reasoning. Identifying these structural gaps motivates our strategy for targeted coverage expansion via clinician-vetted synthetic generation.

\begin{figure}[h!]
    \centering
    \begin{minipage}{0.68\textwidth}
        This expanded coverage is evident in the synthetic MOOVE subset, which shifts towards emergency care settings (from \textbf{15.0\%} in the source data to \textbf{38.7\%} in the synthetic) and life-threatening severities (\textbf{8.6\%} to \textbf{31.8\%}). Similarly, the synthetic Curated QA data significantly alters specialty coverage, notably boosting cardiology (\textbf{3.7\%} to \textbf{32.7\%}) and pulmonology (\textbf{2.9\%} to \textbf{32.2\%}) relative to the source data. It also shifts the age demographic toward adults (from \textbf{35.8\%} to \textbf{84.6\%}) and skews clinical severity away from routine cases (dropping from \textbf{45.6\%} to \textbf{11.7\%}) in favor of urgent (\textbf{28.0\%} to \textbf{67.8\%}) and life-threatening (\textbf{2.2\%} to \textbf{16.3\%}) scenarios. Conversely, the Guidelines dataset maintains a much more stable distribution between its source and synthetic components, consistently emphasizing routine (\textbf{42.4\%} and \textbf{48.2\%}) and urgent (\textbf{39.5\%} and \textbf{41.7\%}) severities within primary and tertiary care levels. Full annotation results are in Appendix~\ref{appendix:data_analysis}.
    \end{minipage}\hfill
    \begin{minipage}{0.3\textwidth}
        \centering
        \includegraphics[width=\textwidth]{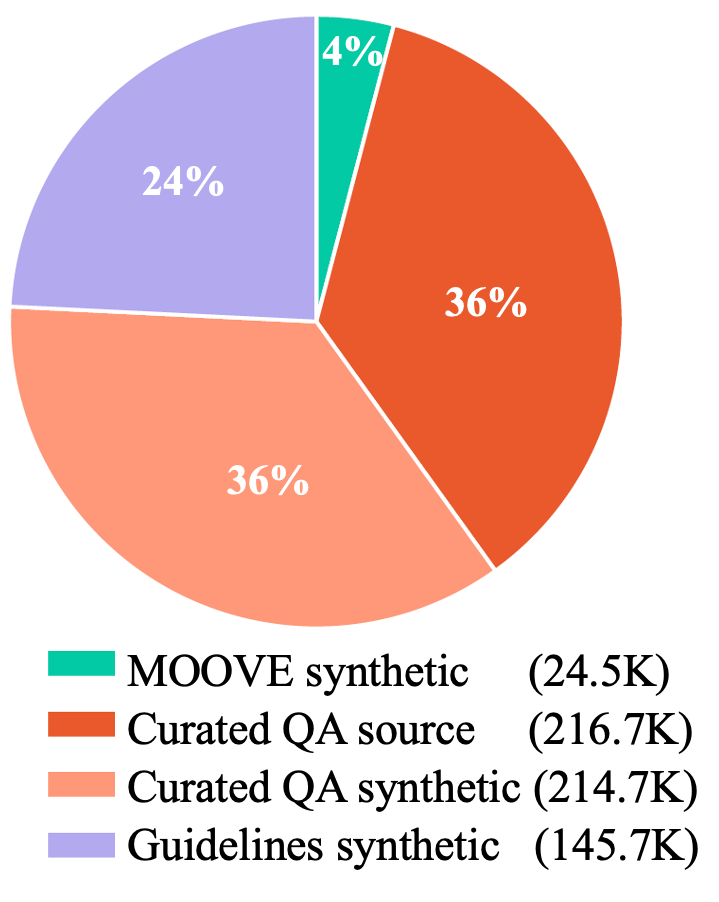}
        \caption{Overview of Fully Open Meditron datasets in records count.}
        \label{fig:data-composition}
    \end{minipage}
\end{figure}

\subsection{Clinician-Vetted Synthetic Coverage Expansion}
To address the identified distributional gaps, we expand the corpus using GPT-OSS-120B to generate targeted synthetic data. Before scaling generation, a panel of four physicians validated the few-shot generation prompts and audited a representative sample of outputs. The panel comprised clinicians with expertise spanning global health, humanitarian response, infectious disease, emergency medicine, primary care, pediatrics and surgery, with clinical experience across Europe, the United States and multiple African settings. For each prompt template, three sampled QA pairs were independently reviewed, with disagreements resolved via panel discussion (prompts in Appendix~\ref{appendix:review_details}).

This review produced four structural improvements to our generation pipeline: (1) refining constraints on "controversial" and "outdated" content to preserve standard-of-care practices in low-resource settings; (2) requiring explicit disease progression and geographic context for epidemiological realism; (3) decoupling stems (which may contain realistic distractors) from answers (which must remain strictly evidence-based); and (4) excluding low-quality evidence sources (e.g. WikiDoc) and overly US-centric phrasing. Following this vetting, we generate three distinct synthetic components:

\noindent \textbullet~ \textbf{Synthetic Curated QA}: Novel exam-style QA pairs seeded from our curated benchmark pool, stratified by question type, incorporating continuous answer-position monitoring to mitigate label bias.

\noindent \textbullet~ \textbf{Guidelines QA}: Question-Answer pairs grounded in 46,469 clinical practice guidelines across 16 global institutions.

\noindent \textbullet~\textbf{Synthetic MOOVE}: Open-ended clinical vignettes seeded from the MOOVE training split to elicit complex diagnostic reasoning.

Synthetic targets are generated using GPT-OSS-120B~\citep{agarwal2025gpt}, selected as the strongest open-source model on the medical training distribution (ablation in table~\ref{tab:judge}). To mitigate hallucinations, labeled examples are rejection-sampled up to eight times at temperature 0.7 until the generated answer matches the gold label under dataset specific regex extraction.

\section{Experimental Setup}

\subsection{Base Models \& Baselines}
We use our corpus for supervised finetuning of five fully open base models: Apertus-70B/8B-Instruct~\citep{apertus2025}, OLMo-2-32B-SFT~\citep{olmo20242}, EuroLLM-22B/9B-Instruct~\citep{ramos2026eurollm}, and one open-weight control, Gemma-3-27B-IT~\citep{gemmateam2025gemma3technicalreport}, to enable a controlled comparison against MedGemma. For each base, we report the unmodified instruction-tuned variant and its MeditronFO finetune under identical decoding settings and prompt templates. Training and code release details are in Appendix~\ref{appendix:training}. In addition to base-vs-finetune comparisons, we report results for three external medical LLMs: MedGemma-27B ~\citep{sellergren2025medgemma}, the strongest open-access medical model trained on undisclosed proprietary data, Llama-3.1-70B-Meditron ~\citep{sallinen2025llama}. For reference on the upper bound of the open-access ecosystem, we also report GPT-OSS-120B ~\citep{agarwal2025gpt}, which is the model used for our synthetic data generation and Qwen3-30B-A3B-Instruct-2507 ~\citep{yang2025qwen3}.

\subsection{Training and Evaluation}
We adapt these base models via supervised fine-tuning on the Fully Open Meditron corpus while preserving each model's native instruction-tuning format. Full training infrastructure, optimizer configurations, and per-model hyperparameters are detailed in Appendix~\ref{appendix:training}.
We evaluate medical knowledge on the test splits of MedQA~\citep{jin2021disease}, MedMCQA~\citep{pal2022medmcqa}, and PubMedQA~\citep{jin2019pubmedqa}, utilizing the held-out MedXpertQA~\citep{zuo2025medxpertqa} as an out-of-distribution check. 
%We also assess open-ended clinical questions using the HealthBench Hard subset \citep{arora2025healthbench} using Qwen3-30B-A3B-Instruct-2507 as judge \citep{yang2025qwen3}. 
As a smoke test that guards against catastrophic forgetting, we evaluate on MMLU-Pro~\citep{wang2024mmlu}, IFEval~\citep{zhou2023instruction}, and ARC-Challenge~\citep{clark2018think}. All evaluations use temperature 0.0; we report accuracy and unweighted averages.

\subsection{Open-Ended Clinical Evaluation}
Standard multiple-choice benchmarks reward structured recall but fail to capture the nuances of open-ended clinical interaction, such as contextual awareness, communication, alignment with guidelines, and harmlessness. We evaluate these dimensions along two axes: 
First, we use the HealthBench evaluation~\citep{arora2025healthbench}, utilizing Qwen3-235B-A22B-Instruct~\citep{yang2025qwen3} as an LLM judge to assess open-ended clinical reasoning against structured, physician-authored rubrics.
Second, we apply Auto-MOOVE, an automated LLM-as-a-judge protocol we developed over clinical prompts drawn from the MOOVE dataset.

For each Auto-MOOVE prompt, two models generate responses, which are passed to our judge models for comparative evaluation. We utilize Qwen3-235B-A22B-Instruct as our primary judge to evaluate the responses and declare an overall winner (Model 1, Model 2, or Tie) and to assign Likert scores from 1 (Poor) to 5 (Excellent) across nine clinical criteria: question comprehension, logical reasoning, relevance and completeness, harmlessness, fairness, contextual awareness, communication, clarity, and alignment with guidelines. Random answer-order swapping is applied at inference to mitigate positional bias, with positions re-mapped during analysis. 

We validate the judge against existing human annotations from MOOVE before using it to evaluate models. Across 204 human raters, the judge's agreement with the panel falls within standard margins of error. Full validation methodology and per-criterion analysis are in Appendix~\ref{appendix:automoove_validation}. To assess sensitivity to judge choice, we additionally report results with GPT-OSS-120B as judge in Section~\ref{sec:results} (Table~\ref{tab:judge}); the qualitative ordering of MeditronFO vs base is preserved across judges for all bases except EuroLLM-22B, where the effect size is smallest. For an overview of the Fully Open Meditron evaluation datasets, refer to Appendix~\ref{appendix:evaluation_datasets}.

%%%%%%%%%%%%%%%%%%%%%%%%%%%%%%%%%%%%%%%%%%%%%%%%%%%%%%%%%%%%
% RESULTS
%%%%%%%%%%%%%%%%%%%%%%%%%%%%%%%%%%%%%%%%%%%%%%%%%%%%%%%%%%%%

\section{Results}
\label{sec:results}

\subsection{Results on Medical QA Benchmarks}

We report comprehensive benchmarking results on established MCQA tasks as well as HealthBench in Table \ref{tab:medbenchtable}. Finetuning a base model on our Fully Open Meditron corpus consistently improves base models. In particular we produce Apertus-70B-MeditronFO, a new state-of-the-art in fully open medical LLMs.

% We evaluate Fully Open Meditron along four axes: open-ended clinical preference, standard medical benchmark performance, general-purpose capability retention, and ablations of the training recipe. Overall, Fully Open Meditron improves every base model on aggregate medical benchmarks and is consistently preferred over its base model in Auto-MOOVE, while revealing tradeoffs between medical specialization and general instruction-following.

\begin{table}[ht]
\centering
\caption{\textbf{Medical benchmark accuracy (\%)}. Every MeditronFO variant improves over its base; gains range from +0.66 (EuroLLM-22B) to +12.80 (Apertus-8B), with smaller bases benefiting most. Apertus-70B-MeditronFO is the strongest fully open model at 53.77 average, narrowing but not closing the gap to MedGemma-27B (60.67). Held-out MedXpertQA tracks the same ordering, indicating gains are not driven by contamination. Best within partition \textbf{bolded}; best fully open \underline{underlined}. For Healthbench, we use the full benchmark and Qwen3-235B-A22B acts as a judge. For a detailed table with older closed open access reference please see appendix \ref{appendix:medical_benchmark_accuracy}}
\label{tab:medbenchtable}
\resizebox{\textwidth}{!}{%
\begin{tabular}{lccccccc}
\toprule
\textbf{Model} & \textbf{MedMCQA} & \textbf{MedQA} & \textbf{PubMedQA} & \textbf{MedXpertQA} & \textbf{Healthbench} & \textbf{Avg} & \textbf{Gain} \\
\midrule
\multicolumn{8}{l}{\textit{Fully open base}} \\
Apertus-70B-Instruct       & 52.43 & 60.64 & 66.80 & 12.33 & 43.72 & 47.18\scriptsize{$\pm$0.84} & --    \\
\quad + \underline{Fully Open Meditron}      & 56.32 & 68.58 & 75.20 & 16.90 & 51.86 & \underline{\textbf{53.77}}\scriptsize{$\pm$0.86} & 6.59  \\
Olmo-2-32B-SFT             & 59.10 & 66.22 & 72.00 & 13.02 & 31.03 & 51.52\scriptsize{$\pm$0.81} & --    \\
\quad + Fully Open Meditron & 57.83 & 69.44 & 76.60 & 17.96 & 44.00 & 53.17\scriptsize{$\pm$0.85} & 1.65  \\
EuroLLM-22B-Instruct       & 54.94 & 66.61 & 73.60 & 14.61 & 43.72 & 50.70\scriptsize{$\pm$0.84} & --    \\
\quad + Fully Open Meditron      & 54.79 & 63.16 & 78.00 & 14.61 & 46.22 & 51.36\scriptsize{$\pm$0.84} & 0.66  \\
\midrule
\multicolumn{8}{l}{\textit{Small Fully open base}} \\
Apertus-8B-Instruct        & 45.80 & 51.14 & 37.60 & 11.71 & 31.61 & 35.57\scriptsize{$\pm$0.76} & --    \\
\quad + Fully Open Meditron      & 48.74 & 58.44 & 75.60 & 13.67 & 45.38 & \textbf{48.37}\scriptsize{$\pm$0.84} & 12.80 \\
EuroLLM-9B-Instruct        & 37.84 & 48.55 & 40.00 & 10.33 & 23.00 & 31.94\scriptsize{$\pm$0.72} & --    \\
\quad + Fully Open Meditron      & 46.98 & 49.73 & 67.40 & 11.63 & 37.53 & 42.65\scriptsize{$\pm$0.81} & 10.71 \\
\midrule
\multicolumn{8}{l}{\textit{Open-access base}} \\
gemma-3-27b-it             & 62.75 & 76.20 & 74.60 & 16.69 & 57.49 & 57.55\scriptsize{$\pm$0.83} & --   \\
\quad + Fully Open Meditron      & 63.71 & 77.61 & 75.80 & 18.00 & 58.02 & 58.63\scriptsize{$\pm$0.83} & 1.08 \\
MedGemma-27B               & 66.44 & 86.10 & 73.00 & 21.88 & 55.92 & \textbf{60.67}\scriptsize{$\pm$0.81} & 3.12 \\
\bottomrule
\end{tabular}}
\end{table}

\subsection{Results on Open-Ended Clinical Evaluation}

\begin{figure}[H]
    \centering
    \includegraphics[width=0.47\textwidth, trim=2cm 0 4.5cm 0, clip]{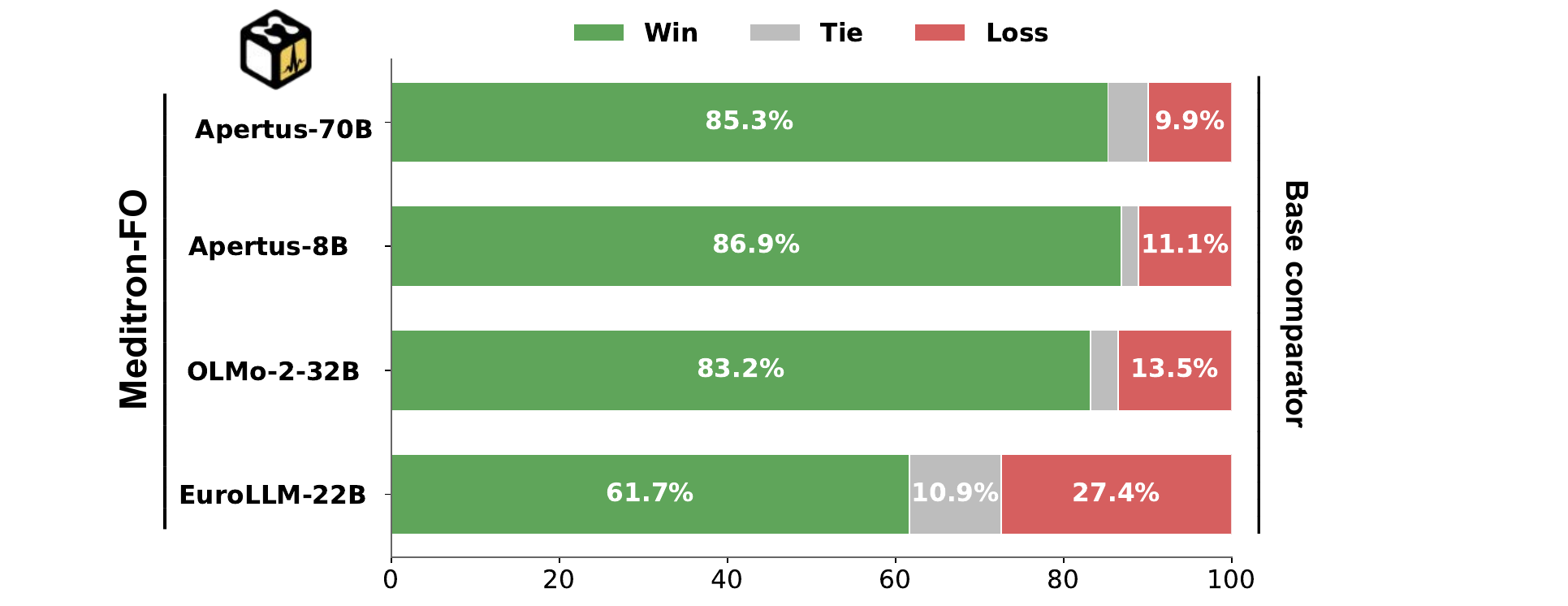}
    \hfill
    \includegraphics[width=0.52\textwidth, trim=1.8cm 0 0 0, clip]{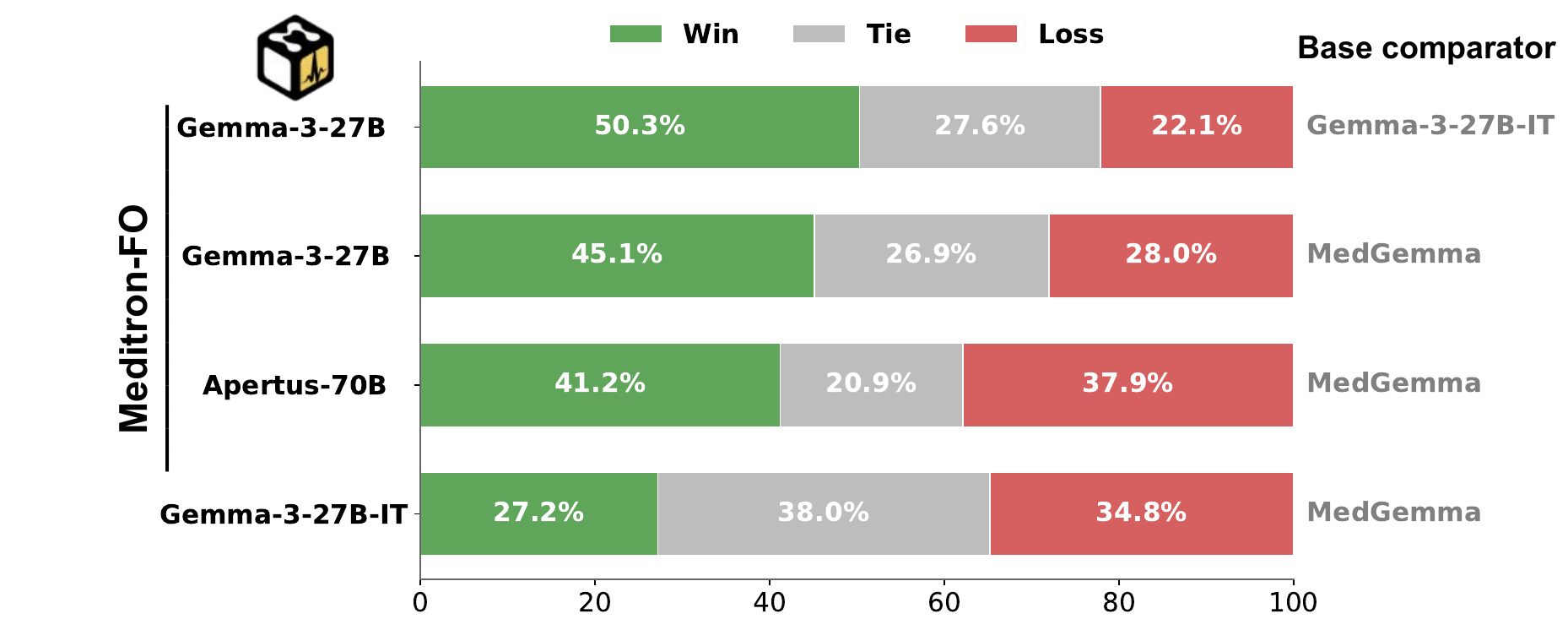}
    \caption{\textbf{Auto-MOOVE pairwise preference results.} For each prompt drawn from the MOOVE evaluation split, two model responses are evaluated by Qwen3-235B-A22B which assigns a winner (Model 1, Model 2, or Tie). Bars show the share of prompts on which each model wins, ties, or loses ($N=12{,}602$ comparisons per pair). Judge agreement with a 204-rater human panel was validated prior to use; see App.~\ref{appendix:automoove_validation}. (Left: Each Fully Open Meditron model versus its corresponding base. Right: Gemma-3-27B-MeditronFO versus MedGemma-27B)}
    \label{fig:auto_moove_1}
\end{figure}

\begin{figure}[h]
    \centering
    \includegraphics[width=0.48\textwidth,trim=0 0 0 2cm, clip]{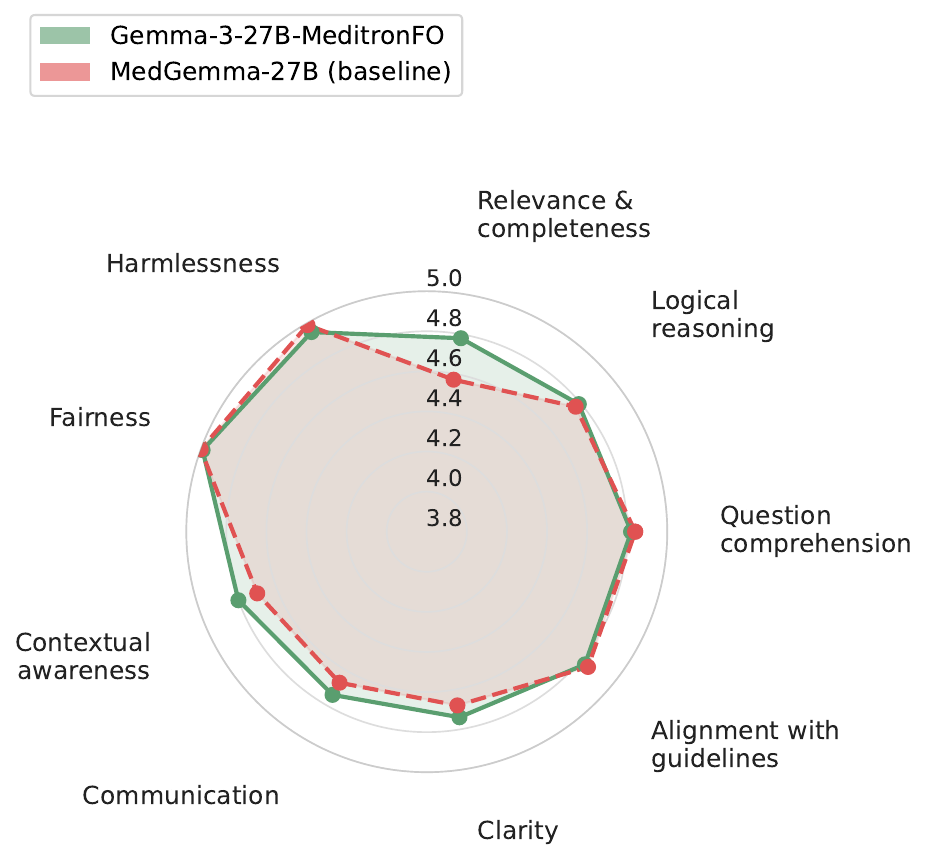}
    \hfill
    \includegraphics[width=0.48\textwidth,trim=0 0 0 2cm, clip]{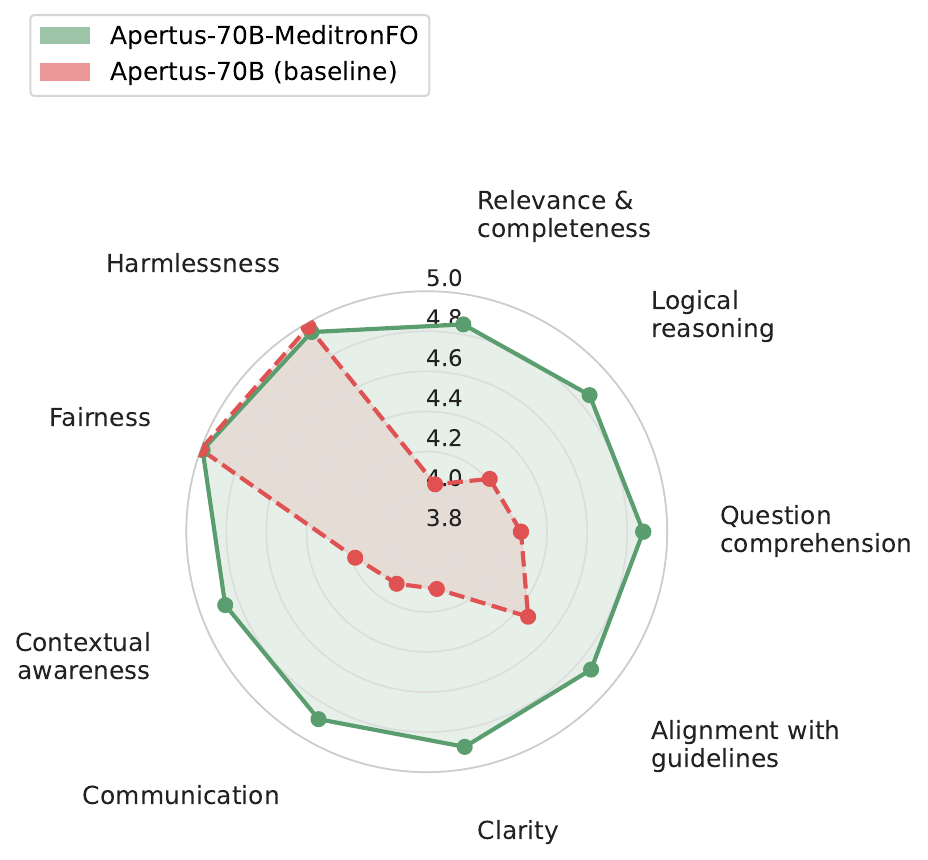}
    \\[1ex]
    \includegraphics[width=0.48\textwidth,trim=0 0 0 2cm, clip]{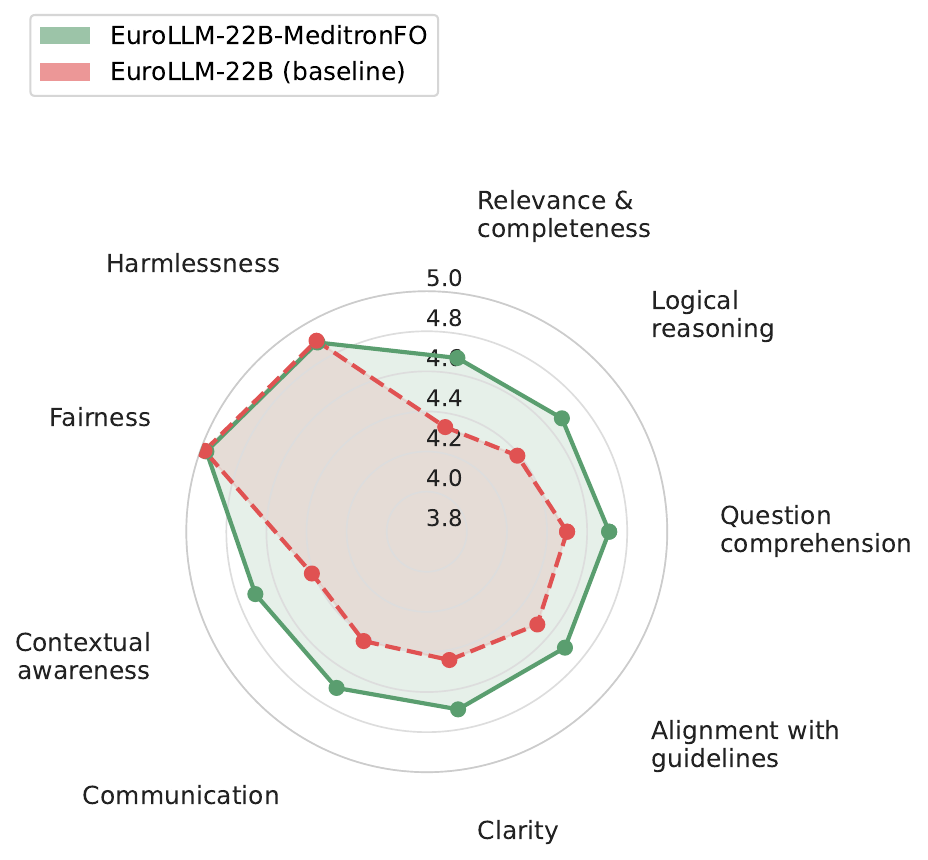}
    \hfill
    \includegraphics[width=0.48\textwidth,trim=0 0 0 2cm, clip]{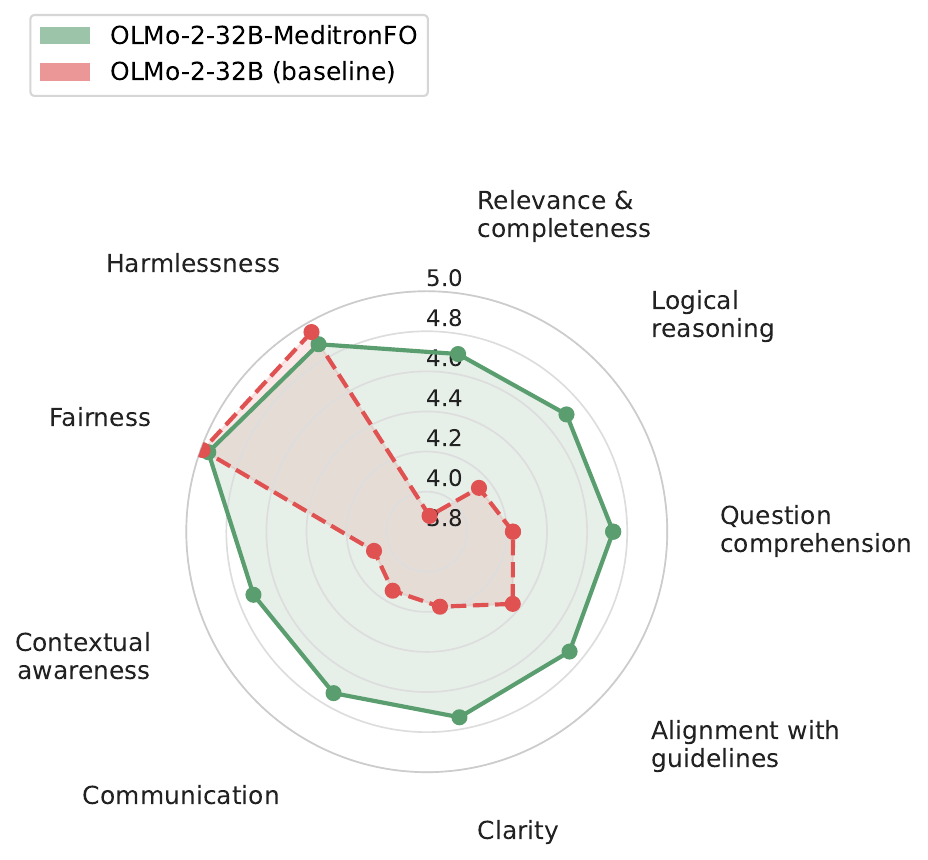}
    \caption{\textbf{Per-criterion Auto-MOOVE Likert profiles for Fully Open Meditron models versus corresponding base.} Panels show (top-left) Gemma 27B, (top-right) Apertus 70B, (bottom-left) EuroLLM 22B, and (bottom-right) OLMo 32B. Axes show mean Likert score (1--5) across the nine evaluation criteria: question comprehension, logical reasoning, relevance and completeness, harmlessness, fairness, contextual awareness, communication, clarity, and alignment with guidelines. Scores are averaged over the same $N=12{,}602$ prompts as Figure~\ref{fig:auto_moove_1} with position-swap debiasing applied, and Qwen3-235B-A22B acts a judge. Larger enclosed area indicates broader improvement across criteria.}
    \label{fig:auto_moove_combined}
\end{figure}

Figure~\ref{fig:auto_moove_1} reports Auto-MOOVE pairwise comparisons between each base model and its Fully Open Meditron finetune. Every *-MeditronFO variant is preferred over its corresponding base, with adjusted win rates ranging from 67.2\% (EuroLLM-22B) to 92\% (Apertus-8B), again with the largest gains observed for smaller bases. Figure~\ref{fig:auto_moove_combined} complements these win-rate results by showing the per-criterion Likert profiles: improvements are not confined to a single axis, but are consistent across clinically relevant dimensions such as question comprehension, logical reasoning, relevance and completeness, contextual awareness, communication, clarity, and alignment with guidelines. EuroLLM-22B shows the smallest margins, consistent with its weaker pairwise preference signal, whereas Apertus-70B, OLMo-2-32B, and Gemma-3-27B exhibit broader gains across criteria. See detailed table of Auto-MOOVE pairwise results in appendix \ref{appendix:auto_moove_results}.

These gains also extend beyond base-versus-finetune comparisons. In cross-model evaluations, Gemma-3-27B-MeditronFO is preferred over MedGemma in 58.6\%, indicating that the improvements are not limited to recovering weaknesses of the underlying bases. This conclusion is further supported by HealthBench (Table~\ref{tab:medbenchtable}), an independent physician-rubric benchmark on which Gemma-3-27B-MeditronFO scores 58.02 compared with 55.92 for MedGemma (+2.1). The agreement between Auto-MOOVE and HealthBench, despite differing judges, prompts, and scoring protocols, argues against the observed gains being a judge-specific or a dataset-distribution artifact.

\subsection{Ablations}
\label{sec:results_ablations}
\textbf{Corpus-component ablations.} The corpus-component ablations in table \ref{tab:corpus_ablations} are the most informative because they identify which parts of the training mixture drive gains on structured medical QA versus open-ended clinical evaluation. The ablations show that there is no single universally optimal recipe: exam-style accuracy, open-ended clinical preference, and general instruction-following pull the training mixture in different directions. The ablation of the Guidelines QA component clearly illustrates this tradeoff. Its removal slightly improves aggregate MCQA accuracy (Med Avg rises from 53.77 to 54.34) while leaving open-ended clinical preference essentially unchanged, consistent with guideline-derived supervision contributing primarily on the margin for exam-style items. Conversely, removing Curated QA produces the largest degradation on both Auto-MOOVE (79.6 drops to 73.4) and $\Delta$  Likert (0.40 to 0.27), indicating that exam-style supervision contributes meaningfully to open-ended clinical quality as well. Removing Synthetic MOOVE also reduces Auto-MOOVE (to 75.5) and $\Delta$  Likert (to 0.34), consistent with its design: vignette-style prompts trade strict exam-format alignment for broader distributional coverage of open-ended diagnostic interaction, i.e. the primary axis measured by Auto-MOOVE.

\begin{table}[h]
\centering
\caption{\textbf{Corpus-component ablations using Apertus-70B as a base.} All runs use identical training settings; only the indicated corpus component is removed. Medical benchmark columns and Medical Avg follow the evaluation protocol of Table 1. Auto-MOOVE reports adjusted win rate under the Qwen3-235B-A22B judge, and ``$\Delta$ Likert'' denotes the mean per-criterion Likert difference averaged across the nine evaluation dimensions. Best values per column are bolded. For Healthbench, we use the full benchmark and Qwen3-235B-A22B acts as a judge. ``Auto-MOOVE'' is the adjusted win rate (\%) of the ablated model against the Apertus-70B-Instruct base under the Qwen3-235B-A22B judge; Extended ablations are in App.~\ref{appendix:ablation_study}.}
\label{tab:corpus_ablations}
\resizebox{\textwidth}{!}{%
\begin{tabular}{lcccccccc}
\toprule
\textbf{Configuration} & \textbf{MedMCQA} & \textbf{MedQA} & \textbf{PubMedQA} & \textbf{HealthBench} & \textbf{MedXpertQA} & \textbf{Med Avg} & \textbf{Auto-MOOVE} & \textbf{$\Delta$ Likert} \\
\midrule
Apertus-70B-Instruct & 52.43 & 60.64 & 66.80 & 43.72 & 12.33 & 47.18 & -- & -- \\
\midrule
\multicolumn{9}{l}{\textit{Corpus component ablations (one component removed at a time)}} \\
Full corpus (default) & 56.32 & 68.58 & 75.20 & 51.86 & 16.90 & 53.77 & \textbf{79.6} & \textbf{0.40} \\
$-$ Curated QA & 54.36 & 65.12 & 71.80 & 40.22 & 17.18 & 49.74 & 73.4 & 0.27 \\
$-$ Synthetic Curated QA & 55.03 & 66.46 & 74.40 & 51.95 & 15.59 & 52.69 & 78.7 & \textbf{0.40} \\
$-$ Guidelines QA & 56.35 & 69.60 & \textbf{76.20} & \textbf{52.22} & 17.35 & \textbf{54.34} & 78.7 & 0.39 \\
$-$ Synthetic MOOVE & \textbf{56.37} & \textbf{69.91} & 72.60 & 51.39 & \textbf{18.20} & 53.69 & 75.5 & 0.34 \\
\bottomrule
\end{tabular}}
\end{table}
We include two additional ablations in Appendix~\ref{appendix:ablation_study}. First, we investigate the retention of general-purpose capacities after fine-tuning on our Fully Open Meditron Corpus, finding only mild drops, which is in line with other domain-specific fine-tuning recipes. Additionally, we analyze a range of different judge models for Auto-MOOVE, finding a consistent preference for the MeditronFO variants over their corresponding base models across judges.

\section{Discussion}

Evaluating clinical LLMs requires moving beyond traditional multiple-choice question answering (MCQA) to assess genuine clinical interaction. After finetuning on the Fully Open Meditron Corpus, Apertus-70B-MeditronFO establishes a new state of the art among fully open medical models. It achieves strong performance on both MCQA benchmarks and LLM-as-a-judge evaluations including Auto-MOOVE and HealthBench.

Importantly, these gains generalize across model families: every finetuned model improves over its base in both structured and open-ended evaluation. This supports the central premise of the fully open paradigm: clinically competitive medical specialization can be achieved using reproducible, auditable pipelines rather than opaque adaptation procedures. The resulting corpus provides a reusable foundation for training and evaluating future fully open medical models. Notably, Gemma-27B-MeditronFO surpasses MedGemma-27B on both HealthBench and Auto-MOOVE despite being derived from a fully open pipeline.

% \section{Discussion}

% \textbf{Conclusion.} The benchmark-driven supervised fine-tuning recipes that dominate open medical LLM development can be beneficially augmented by a structured corpus-construction pipeline that combines aggregated public benchmarks with clinician-vetted synthetic components covering exam-style QA, guideline-grounded QA, and clinical vignettes. The Fully Open Meditron framework provides a way to translate clinical evaluation desiderata—contextual awareness, communication, alignment with guidelines, and harmlessness—into actionable training signal for contemporary open base models. At the same time, fully open base models address a weakness of closed-data specialists, providing a way to build medical LLMs whose training data, generation prompts, and evaluation protocols are end-to-end auditable, which is what clinicians, regulators, and patients have a legitimate interest in inspecting. We see this intersection of structured corpus construction with fully open foundation models as an exciting direction for advancing the accessibility, auditability, and clinical relevance of medical AI.

\textbf{Limitations and future directions.} 
Several specific limitations warrant attention: Auto-MOOVE judge agreement falls below the median human rater and is systematically less discriminating than clinicians on safety-relevant criteria such as harmlessness and fairness, making it unsuitable as a deployment-readiness signal for these dimensions; our decontamination is syntactic rather than semantic, leaving open the possibility that a teacher paraphrases or generalizes evaluation content when seeded from the corresponding training split; instruction-following degrades on some bases, suggesting the uniform 10\% Tülu replay should be tuned per base; synthetic data accounts for roughly 64\% of the corpus while clinician auditing covered only three sampled QA pairs per generation prompt template, bounding systematic but not item-level errors; and a single teacher (GPT-OSS-120B) and single judge introduce model-specific stylistic and reasoning biases that our ablations probe but do not eliminate. Finally, this work focuses on supervised fine-tuning of off-the-shelf bases, and incorporating preference optimization, continued pretraining on the GUIDELINES corpus, or end-to-end open-provenance teachers might present opportunities to further enhance the auditability and clinical capabilities of fully open medical LLMs.

\subsection{Broader impact}

Fully Open Meditron is intended to advance the auditability of medical AI by making the full training pipeline inspectable. The accompanying risks are those general to medical LLMs: confidently incorrect outputs, propagation of training-data biases, and misuse as a substitute for clinical judgment. The fact that the corpus is open is a partial mitigation (it enables third-party auditing and red-teaming) and a partial amplifier (the recipe is reproducible by parties who may not perform equivalent audits). We release the corpus under a research-use license and recommend that downstream practitioners conduct domain-specific safety evaluation before any deployment-adjacent use.

%%%%%%%%%%%%%%%%%%%%%%%%%%%%%%%%%%%%%%%%%%%%%%%%%%%%%%%%%%%%
% ACK & BIBLIOGRAPHY
%%%%%%%%%%%%%%%%%%%%%%%%%%%%%%%%%%%%%%%%%%%%%%%%%%%%%%%%%%%%

\begin{ack}
This work was supported under project ID \#27 as part of the Swiss AI Initiative, through a grant from the ETH Domain and computational resources provided by the Swiss National Supercomputing Centre (CSCS) under the Alps infrastructure.

We thank the physician review panel within the LiGHT laboratory for their clinical auditing, methodological review, and validation of the synthetic generation and evaluation pipelines. We additionally acknowledge the many physicians and clinical experts who contributed to the MOOVE initiative through expert review, pairwise evaluation, benchmarking, and clinical vignette development across diverse international settings.
\end{ack}

\bibliographystyle{unsrt} % Or 'unsrt' if you want them in order of appearance
\bibliography{references} 

%%%%%%%%%%%%%%%%%%%%%%%%%%%%%%%%%%%%%%%%%%%%%%%%%%%%%%%%%%%%
% APPENDIX
%%%%%%%%%%%%%%%%%%%%%%%%%%%%%%%%%%%%%%%%%%%%%%%%%%%%%%%%%%%%

\appendix

% --- Color definitions ---
\definecolor{promptbg}{HTML}{EAF1FB}
\definecolor{medgemmabg}{HTML}{F5F5F5}
\definecolor{meditronbg}{HTML}{EAF6EC}
\definecolor{judgebg}{HTML}{FFF8E1}
\definecolor{promptrule}{HTML}{4A78B8}
\definecolor{medgemmarule}{HTML}{888888}
\definecolor{meditronrule}{HTML}{3B8B5A}
\definecolor{judgerule}{HTML}{C9A227}

% --- Reusable box styles ---
\newtcolorbox{promptbox}[1][]{
  enhanced, breakable, colback=promptbg, colframe=promptrule,
  boxrule=0.4pt, arc=2pt, left=8pt, right=8pt, top=6pt, bottom=6pt,
  title=\textbf{Prompt}, fonttitle=\bfseries\small,
  coltitle=promptrule, colbacktitle=promptbg,
  attach boxed title to top left={xshift=6pt, yshift=-7pt},
  boxed title style={colback=promptbg, colframe=promptbg, sharp corners},
  #1
}

\newtcolorbox{medgemmabox}[1][]{
  enhanced, breakable, colback=medgemmabg, colframe=medgemmarule,
  boxrule=0.4pt, arc=2pt, left=8pt, right=8pt, top=6pt, bottom=6pt,
  title=\textbf{MedGemma-27B}, fonttitle=\bfseries\small,
  coltitle=black, colbacktitle=medgemmabg,
  attach boxed title to top left={xshift=6pt, yshift=-7pt},
  boxed title style={colback=medgemmabg, colframe=medgemmabg, sharp corners},
  #1
}

\newtcolorbox{meditronbox}[1][]{
  enhanced, breakable, colback=meditronbg, colframe=meditronrule,
  boxrule=0.4pt, arc=2pt, left=8pt, right=8pt, top=6pt, bottom=6pt,
  title=\textbf{Gemma-Meditron (ours)}, fonttitle=\bfseries\small,
  coltitle=meditronrule, colbacktitle=meditronbg,
  attach boxed title to top left={xshift=6pt, yshift=-7pt},
  boxed title style={colback=meditronbg, colframe=meditronbg, sharp corners},
  #1
}

\newtcolorbox{judgebox}[1][]{
  enhanced, breakable, colback=judgebg, colframe=judgerule,
  boxrule=0.4pt, arc=2pt, left=8pt, right=8pt, top=6pt, bottom=6pt,
  title=\textbf{Judge Verdict}, fonttitle=\bfseries\small,
  coltitle=black!70, colbacktitle=judgebg,
  attach boxed title to top left={xshift=6pt, yshift=-7pt},
  boxed title style={colback=judgebg, colframe=judgebg, sharp corners},
  #1
}

\section{Examples where Gemma-Meditron wins against MedGemma}
\label{sec:gemma_meditron_wins}

We present two qualitative examples where Gemma-Meditron is preferred over MedGemma by an LLM-as-a-judge evaluation. and the judge's preference reflects substantive differences in clinical reasoning, contextual awareness, and structured presentation.

% =====================================================================
\subsection*{Example 1}
% =====================================================================

\begin{promptbox}
\small
I am the emergency radiologist at the CHUV in Lausanne who has been asked to perform imaging in an 18-year-old man transported to the hospital after a severe bicycle accident. He is conscient and his Glasgow Coma Scale is 15, but hemodynamically, he is quite unstable. Furthermore, he suffers from severe upper abdominal pain. What imaging modality and what acquisition protocol do you recommend? What is the most likely diagnosis?
\end{promptbox}

\vspace{4pt}

\begin{medgemmabox}
\small
Okay, this is a critical situation. An 18-year-old male, high-energy trauma (bicycle accident), GCS 15 (good neurological status), but hemodynamically unstable with severe upper abdominal pain. This immediately raises concerns for significant intra-abdominal hemorrhage, likely from solid organ injury (liver, spleen) or mesenteric injury.

Here's my recommendation for imaging modality and protocol, along with the most likely diagnosis:

\textbf{1. Imaging Modality: Contrast-Enhanced CT Scan}

*   \textbf{Why CT?} CT is the gold standard for evaluating hemodynamically unstable trauma patients with suspected abdominal or thoracic injury. It is fast, widely available, and provides detailed anatomical information about solid organs, vasculature, retroperitoneum, and bony structures.
*   \textbf{Why Contrast Enhanced?} While a non-contrast CT can identify free fluid (hemoperitoneum), contrast enhancement is crucial for:
    *   Identifying active arterial extravasation (bleeding).
    *   Assessing the extent and grade of solid organ injuries.
    *   Evaluating vascular injuries (e.g., aorta, vena cava, mesenteric vessels).
    *   Visualizing bowel and mesenteric injuries.
    *   Differentiating between solid organ hematomas and active bleeding.

\textbf{2. Acquisition Protocol: Trauma CT Protocol (Optimized for Hemodynamic Instability)}

Given the patient's hemodynamic instability, speed and efficiency are paramount. The protocol should be streamlined and focused.

*   \textbf{Scanner:} Use a high-speed, multi-detector CT scanner (64-slice or higher preferred).
*   \textbf{Patient Positioning:} Supine position with arms above the head.
*   \textbf{Scanning Range:}
    \emph{   \textbf{Head:} Non-contrast CT head (brief, to rule out intracranial injury despite GCS 15, as focal neurological deficits can be subtle or develop later). }This can potentially be skipped or deferred if the patient is extremely unstable and abdominal/thoracic injury is the primary concern, but it's generally standard in trauma.*
    *   \textbf{Neck:} Non-contrast CT neck (to evaluate for cervical spine injury, especially with a bicycle accident).
    *   \textbf{Chest:} Contrast-enhanced CT chest (thin slices, 1-1.5mm). Essential to rule out hemothorax, pneumothorax, pulmonary contusion, aortic injury, and mediastinal hematoma.
    *   \textbf{Abdomen \& Pelvis:} Contrast-enhanced CT abdomen and pelvis (thin slices, 1-1.5mm). This is the most critical part for this patient.
*   \textbf{Contrast Administration:}
    *   \textbf{Type:} Iodinated intravenous contrast (e.g., iohexol, iopamidol).
    *   \textbf{Volume:} Sufficient volume to adequately opacify the aorta and hepatic veins (e.g., 100-150 mL in an adult, adjusted for weight).
    *   \textbf{Injection Rate:} Relatively high rate (e.g., 4-5 mL/sec) to achieve rapid arterial phase enhancement.
    \emph{   \textbf{Timing:} Use a }dual-phase\emph{ or }tri-phase* protocol if feasible, but prioritize the arterial phase.
        *   \textbf{Arterial Phase:} Essential for detecting active bleeding and assessing vascular injury (typically 20-30 seconds post-injection).
        *   \textbf{Portal Venous Phase:} Helps delineate venous injuries and assess organ perfusion (typically 60-80 seconds post-injection).
        *   \textbf{Delayed Phase (Optional/If Time Permits):} Can be useful for identifying urinary tract injuries, but may be skipped in an unstable patient to save time.
*   \textbf{Reconstruction:}
    *   Axial images.
    *   Coronal and sagittal reformations (especially for abdomen/pelvis).
    *   Thin-section multiplanar reconstructions (MPR) as needed.
*   \textbf{Speed:} Maximize scan speed while maintaining diagnostic quality. Instruct the CT technologist to minimize delays between phases.
*   \textbf{Monitoring:} Continuous monitoring of vital signs during the scan. Have resuscitation equipment readily available. Inform the resuscitation team of the scan progress.

\textbf{3. Most Likely Diagnosis:}

Given the mechanism (bicycle accident), hemodynamic instability, and severe upper abdominal pain, the most likely diagnoses involve significant \textbf{intra-abdominal hemorrhage}.

*   \textbf{Splenic Injury:} The spleen is the most commonly injured solid organ in blunt abdominal trauma. Rupture or laceration can lead to massive hemoperitoneum and shock.
*   \textbf{Liver Injury:} The liver is the second most commonly injured organ. Lacerations, hematomas, or vascular injuries (e.g., hepatic artery, portal vein) can cause significant bleeding.
*   \textbf{Mesenteric Injury:} Injury to the mesentery or its vessels can cause significant venous or arterial bleeding
\end{medgemmabox}

\vspace{4pt}

\begin{meditronbox}
\small
Okay, this is a classic trauma scenario. A young, unstable patient with upper abdominal pain after a high-impact mechanism (bicycle accident) demands a rapid, comprehensive imaging workup. Here's my recommendation for imaging modality, protocol, the most likely diagnosis, and a brief discussion of why this approach is chosen, tailored to the situation at CHUV (a high-volume trauma center).

\textbf{1. Imaging Modality: Contrast-Enhanced Multidetector CT (MDCT) of the Abdomen and Pelvis -- FAST-CT Protocol}

\emph{   \textbf{Why CT, not US?} While bedside ultrasound (FAST exam) is }always\emph{ first-line in the ED for the "ABCDE" assessment and to detect free fluid, in a patient who is already being taken for definitive imaging, and who has }specific* abdominal pain, a CT offers a far more detailed and complete evaluation of solid organ injury, vascular compromise, bowel perforation, and pelvic fracture.  The patient's GCS of 15 allows for a quick, whole-body CT scan.
*   \textbf{Why contrast?}  Contrast is essential for evaluating liver, spleen, kidney, and bowel perfusion, detecting active bleeding, and differentiating solid organ injury from hematoma. In a hemodynamically unstable patient, contrast helps identify sources of ongoing hemorrhage.
*   \textbf{Why 'FAST-CT'?} This is a tailored protocol specifically designed for trauma. It prioritizes speed and efficiency while maximizing diagnostic yield.

\textbf{2. Acquisition Protocol (FAST-CT -- Adapted for CHUV resources)}

| Parameter | Recommended Setting (CHUV-typical) | Rationale |
|-----------|-------------------------------------|-----------|
| \textbf{Scanner} | 128- or 256-slice (or higher) | Faster rotation, thinner slices |
| \textbf{Gantry Tilt} | 0-15° (depending on patient habitus) | Improves visualization of the upper abdomen |
| \textbf{Inspiration Breath-Hold} | 10-15 seconds (if possible) | Reduces motion artifact; if patient cannot hold breath, use high-pitch spiral with respiratory gating |
| \textbf{Scan Range} | Upper dome of the liver to pubic symphysis (full abdomen \& pelvis) | Captures all potential injury sites |
| \textbf{Slice Thickness} | 2.5-3 mm (reconstructed to 1.25 mm) | Adequate detail for most injuries; thinner reconstructions for vascular imaging |
| \textbf{Contrast Protocol (IV Dual-energy Acquisition)} |  - \textbf{Arterial Phase (30-40 s delay)}: 1.5-2 mL/kg iodinated contrast (e.g., Iomeprol 400 mg/mL) at 4-5 mL/s flow rate. - \textbf{Portal Venous Phase (60-90 s delay)}:  Repeat bolus of 0.8-1 mL/kg same contrast agent. - \textbf{Delayed Phase (5-10 min)}: Can be added if suspected bowel injury or delayed hemorrhage; 0.5 mL/kg |  Arterial phase shows active bleeding, portal venous phase defines parenchymal lacerations, delayed phase identifies bowel leak. |
| \textbf{Reconstruction Algorithms} | - \textbf{Standard Reconstruction:} Soft-tissue window (B30f/45s) for parenchyma. - \textbf{Vascular Reconstruction:} Thin-slice MIP/VR for aorta, IVC, SMV, portal vein, hepatic arteries. - \textbf{Bone Windows:} For pelvic fracture assessment. |  Optimizes visualization of different tissues. |
| \textbf{Radiation Dose Reduction} | Use automated tube current modulation (CARE Dose 4D/kV-switching) and iterative reconstruction (e.g., iDose4) | Minimizes radiation exposure without sacrificing image quality. |
| \textbf{Image Post-processing} | - \textbf{Volume Rendering (VR/MIP)}: For vascular assessment. - \textbf{Multiplanar Reconstructions (MPR)}: Axial, coronal, sagittal planes. - \textbf{3D Segmentation (if available)}: For quantifying hematoma volume. | Facilitates surgical planning. |
| \textbf{Total Acquisition Time} | \textasciitilde{}10-15 seconds for the dynamic phases (arterial, portal) + 15-20 seconds for the delayed phase. Total scan time <40-60 seconds. | Critical in unstable patients. |
| \textbf{Image Reporting} |  Structured reporting using the AAST-OIS injury grading system for each organ (liver, spleen, pancreas, kidney, bowel, mesentery, adrenal glands). |  Standardized communication with surgeons. |

\textbf{Important Protocol Considerations at CHUV:}

*   \textbf{Dual-energy CT (DECT):} CHUV has advanced DECT capabilities.  Use DECT to perform virtual unenhanced images and iodine maps. This can reduce contrast-induced
\end{meditronbox}

\vspace{4pt}

\begin{judgebox}
\small
Both models demonstrate excellent understanding of the clinical scenario, recognizing the urgency of hemodynamic instability, the need for rapid imaging, and the high likelihood of intra-abdominal hemorrhage following blunt trauma. Model 1 provides a clear, accurate, and clinically sound recommendation for contrast-enhanced CT with appropriate protocol details, including phase timing, contrast administration, and reconstruction. It correctly identifies splenic and hepatic injuries as the most likely diagnoses. Model 2 builds upon this with superior contextual awareness---specifically referencing CHUV's advanced capabilities (e.g., dual-energy CT, DECT, iDose4, 3D segmentation), which enhances relevance and realism for the setting. It also includes a structured table that improves clarity and clinical utility, and it integrates radiation dose optimization and structured reporting, aligning with current trauma imaging best practices. While Model 1 is thorough and correct, Model 2 offers more nuanced, institutionally tailored, and technically advanced guidance without sacrificing clarity or safety. Both are harmless, fair, and well-aligned with guidelines. Model 2's contextual specificity and enhanced technical detail give it a slight edge.
\end{judgebox}

% =====================================================================
\subsection*{Example 2}
% =====================================================================

\begin{promptbox}
\small
I am a pediatric intensive care physician working at CHUV in Lausanne in Switzerland. I take care of a 14-year-old female patient following a severe traumatic brain injury with diffuse cerebral edema and multiple parenchymal hemorrhages on head CT. What is the use of dosing biomarkers to better assess her prognosis ?
\end{promptbox}

\vspace{4pt}

\begin{medgemmabox}
\small
Okay, this is a critical situation, and using biomarkers is becoming increasingly important in TBI management, especially in severe cases like the one you're describing. Here's a breakdown of how dosing biomarkers can be used to better assess prognosis in a 14-year-old female patient with severe TBI, diffuse cerebral edema, and multiple parenchymal hemorrhages at CHUV:

\textbf{Understanding the Goal: Why Biomarkers?}

Traditional prognostic tools in TBI (like GCS, age, pupillary response, CT findings) are valuable but have limitations. They provide a snapshot in time and don't capture the complex, dynamic biological processes occurring after injury. Biomarkers aim to:

1.  \textbf{Detect Injury Severity More Accurately:} Go beyond macroscopic damage seen on CT.
2.  \textbf{Track Pathophysiological Processes:} Monitor inflammation, neuronal injury, and metabolic changes.
3.  \textbf{Predict Long-Term Outcomes:} Help differentiate patients likely to recover well from those with a high risk of poor outcomes (death, severe disability).
4.  \textbf{Guide Management:} Potentially inform therapeutic decisions (e.g., intensity of monitoring, interventions to manage secondary injury).
5.  \textbf{Provide Earlier Prognostication:} Offer insights sooner than clinical assessments alone, which can be crucial for family counseling and resource allocation.

\textbf{Key Biomarkers for Severe TBI Prognosis (and their relevance to your patient):}

Several biomarkers are being investigated and increasingly used, often in combination. Here are the most relevant ones:

1.  \textbf{Neuron-Specific Enolase (NSE):}
    *   \textbf{What it is:} A glycolytic enzyme found primarily in neurons. Released into CSF and blood upon neuronal damage/death.
    *   \textbf{Relevance:} Elevated levels correlate with the extent of neuronal injury. Very high levels early on are associated with poor prognosis (death or severe disability).
    *   \textbf{Use:} Often measured in serum or CSF within the first 24-72 hours. Serial measurements can track the trajectory of injury. In your patient with diffuse edema and hemorrhages, NSE is likely to be significantly elevated, and its level is a strong prognostic indicator.

2.  \textbf{S100B (Soluble Protein 100B):}
    *   \textbf{What it is:} A calcium-binding protein, primarily expressed by astrocytes. Released into CSF and blood after brain injury, particularly involving glial cells.
    *   \textbf{Relevance:} Reflects astrocytic damage and blood-brain barrier disruption. Also associated with injury severity and outcome. While less specific to neurons than NSE, it's often elevated in TBI.
    \emph{   \textbf{Use:} Measured in serum or CSF, often within the first 24-48 hours. High levels are associated with worse outcomes. Its utility might be slightly less pronounced than NSE for }severe* TBI prognosis, but it's still valuable, especially when combined with other markers.

3.  \textbf{Glial Fibrillary Acidic Protein (GFAP):}
    *   \textbf{What it is:} An intermediate filament protein specific to astrocytes. Released into CSF and blood following astrocyte injury.
    *   \textbf{Relevance:} Increasingly recognized as a sensitive and relatively specific marker of brain injury, particularly astrocytic damage. It appears to be more stable in serum than S100B.
    *   \textbf{Use:} Measured in serum or CSF. Emerging evidence suggests GFAP may be a powerful prognostic marker, potentially even more so than NSE or S100B in some studies, especially when used in combination or as part of panels. Its availability might be more limited depending on the lab, but it's a key marker to consider.

4.  \textbf{Ubiquitin C-terminal Hydrolase L1 (UCH-L1):}
    *   \textbf{What it is:} An enzyme involved in protein degradation, found predominantly in neurons. Released into CSF and blood after neuronal injury.
    *   \textbf{Relevance:} Similar to NSE, it indicates neuronal damage. Some studies suggest it might be more specific to axonal injury than NSE.
    *   \textbf{Use:} Measured in serum or CSF. High levels correlate with injury severity and poor outcome. Often used in conjunction with other markers.

5.  \textbf{Neurofilament Light Chain (NfL):}
    *   \textbf{What it is:} A structural protein of neuronal axons. Released into CSF and blood when axons are damaged or degenerate.
    *   \textbf{Relevance:} A marker of axonal injury. Elevated levels are associated with TBI severity and poorer outcomes. It's also being investigated for tracking recovery and neurodegeneration.
    *   \textbf{Use:} Measured in serum or CSF. Its role in acute TBI prognosis is still being refined, but it shows promise, particularly for detecting subtle or evolving axonal damage.

\end{medgemmabox}

\vspace{4pt}

\begin{meditronbox}
\small
\textbf{Background}

Traumatic brain injury (TBI) in the pediatric population is a major cause of morbidity and mortality, and timely identification of patients at high risk for severe outcomes is critical for resource allocation and shared decision-making with families.  Conventional imaging (CT, MRI) and clinical scores (e.g., Glasgow Coma Scale, Pediatric Glasgow Outcome Scale) are often insufficient for precise prognostication, especially in the acute phase.

\textbf{Emerging Role of Biomarkers in Pediatric TBI}

Biomarkers can provide a window into the neurobiological response to injury, offering objective data that complements clinical assessment.  They fall into several categories, some more established than others:

| \textbf{Biomarker Class} | \textbf{Examples} | \textbf{Timing of Measurement (from injury)} | \textbf{Physiologic Rationale} | \textbf{Prognostic Value (pediatric TBI)} | \textbf{Practical Considerations for CHUV} |
|----------------------|--------------|----------------------------------------|---------------------------|-------------------------------------|-----------------------------------|
| \textbf{Neuronal Injury} | \textbf{GFAP (Glial Fibrillary Acidic Protein)} | \$\textbackslash\{\}leq\$ 12-24 hours   | Marker of astrocyte injury, correlates with lesion volume | Strong predictor of unfavorable outcome (death, severe disability) at 6 months.  High sensitivity, specificity in pediatric data. | Point-of-care assay available; can be ordered from the central lab if not emergent. |
| | \textbf{Uch-L1 (Ubiquitin C-terminal hydrolase-L1)} | \$\textbackslash\{\}leq\$ 12-24h | Cytosolic protein released from neurons/glia upon injury | Adds prognostic information when combined with GFAP; predicts long-term disability. | Standard laboratory panel, readily available. |
| \textbf{Astrocyte activation/astroglial injury} | \textbf{S100B} | 24-72h | Released from astrocytes; can be non-specific (e.g., also elevated in cardiac injury). |  Less robust than GFAP/Uch-L1, especially in moderate TBI.  May be useful as part of a multi-marker panel. | Already measured in routine CHUV ICU labs. |
| \textbf{Axonal Injury} | \textbf{NF-L (Neurofilament Light Chain)} | 24-72h, then serial measurements  | Released from damaged axons; reflects neuronal degeneration. | Emerging as a strong prognosticator, especially for long-term neurocognitive deficits. | Not yet standard at CHUV; requires a send-out assay or local validation. |
| \textbf{Inflammation/Immune Response} | \textbf{IL-6 (Interleukin-6), TNF-\$\textbackslash\{\}alpha\$, HMGB1} | 6-24h, then 24-72h | Reflect early inflammatory cascade; correlate with secondary injury. | Predictive of cerebral edema, ICU length of stay, and functional outcome, but less specific than neuronal markers. | CHUV immunology lab can measure these. |
| \textbf{Metabolic/Cerebral Energy Crisis} | **Plasma glucose, lactate, pyruvate, cerebral microdialysis (lactate/pyruvate ratio)  | Continuous monitoring (lactate/pyruvate) or spot samples. | Early marker of mitochondrial dysfunction and energy failure. | Useful for guiding resuscitation (e.g., glycemic control, oxygen delivery) but limited as a standalone prognosticator. | Standard ICU monitoring; microdialysis not routinely used at CHUV. |

\textbf{How to Use Biomarkers in Your 14-Year-Old Patient at CHUV}

1. \textbf{Acute Phase (first 24-48 hours)}
   - \textbf{GFAP + Uch-L1:} This combination is the most evidence-based for early prognostication in pediatric severe TBI. Order them stat from the CHUV central lab.  
   - \textbf{Interpretation:} 
     - \textbf{High GFAP (> 200-400 pg/mL) + high Uch-L1 (> 50 ng/mL)}:  Strongly predicts an increased risk of death or severe disability at 6 months (based on the Traumatic Brain Injury Predictive Study).  This information helps frame the discussion with the family about potential outcomes and the appropriateness of aggressive interventions (e.g., prolonged mechanical ventilation, neurocritical care).
     - \textbf{Discordant results (e.g., high GFAP, normal Uch-L1)}:  Suggests astrocyte injury may be present but neuronal injury is less severe.  May warrant a more expectant approach while continuing intensive monitoring.
   - \textbf{S100B:}  May be obtained simultaneously as it is already measured in your ICU panel, but its prognostic value is lower when GFAP/Uch-L1 are available.

2. \textbf{Subacute Phase (24-72 hours)}
   - \textbf{NF-L:} If the
\end{meditronbox}

\vspace{4pt}

\begin{judgebox}
\small
\textit{Note: in the raw judge output, the labels ``Model 1'' and ``Model 2'' are swapped relative to the model identities shown above (the evaluation randomizes presentation order to control for position bias).}

Model 1 demonstrates superior clinical reasoning, depth of knowledge, and structured presentation. It accurately identifies the most relevant biomarkers for pediatric severe TBI (GFAP, Uch-L1, NF-L, S100B, inflammatory markers), provides precise timing, physiological rationale, prognostic value, and practical implementation guidance tailored to CHUV's lab capabilities. The use of a table enhances clarity and facilitates rapid clinical decision-making. It also includes nuanced interpretation of biomarker combinations and their implications for family counseling and treatment intensity---critical in a complex case like this.  

Model 2 offers a reasonable overview but lacks depth and specificity. It omits key biomarkers (e.g., NF-L, IL-6, lactate/pyruvate ratio) and fails to integrate contextual details about CHUV's lab infrastructure. The response is cut off mid-sentence, indicating incompleteness. While it correctly identifies NSE, S100B, GFAP, Uch-L1, and NfL, it does not provide actionable guidance on timing, interpretation thresholds, or integration with clinical and imaging data. The communication is less structured, and the lack of a table or clear decision framework reduces utility in a high-acuity setting.  

Both models are harmless and fair, but Model 1 is significantly more aligned with clinical guidelines for neurocritical care and evidence-based prognostication in pediatric TBI.
\end{judgebox}

\newpage
\section{Data analysis}
\label{appendix:data_analysis}

The training mix combines a curated pool of public medical QA with three
synthetic components seeded from real corpora
(Table~\ref{tab:training_datasets}): exam-style QA seeded from the curated
pool(Table~\ref{tab:training_datasets_curated}), QA grounded in clinical practice guidelines, and
clinical-vignette prompts seeded from MOOVE training data. Synthetic data
accounts for $\sim\!64\%$ of examples and $\sim\!71\%$ of tokens,
motivating the source-versus-synthetic distribution checks reported below.

\begin{table}[H]
\centering
\caption{Overview of Fully Open Meditron datasets}
\label{tab:training_datasets}
\small
\resizebox{\textwidth}{!}{%
\begin{tabular}{l l r r l}
\toprule
\textbf{Component} & \textbf{Dataset} & \textbf{No. examples} & \textbf{Tokens} & \textbf{Description} \\
\midrule
Curated QA & 8 medical QA datasets & 216{,}719 & 43.9M & Aggregated public medical QA (training splits) \\ 
\midrule
\multirow{3}{*}{Synthetic}
 & Synthetic Curated QA & 214{,}654 & 70.4M & Novel exam-style QA seeded from Curated QA pool \\
 & Guidelines QA        & 145{,}681 & 12.5M & QA grounded in 46{,}469 clinical practice guidelines \\
 & Synthetic MOOVE      & 24{,}465  & 23M & Clinical vignette prompts seeded from MOOVE training \\
%\midrule
%Replay & Tülu 3 SFT (10\% mix) & 93{,}934 & TODO & General instruction-following replay \\
\midrule
\textbf{Total} & & \textbf{601{,}519} & \textbf{150M} & \\
\bottomrule
\end{tabular}}
\end{table}

\begin{figure}[h]
  \centering
  \begin{subfigure}[t]{0.95\textwidth}
    \includegraphics[width=\linewidth]{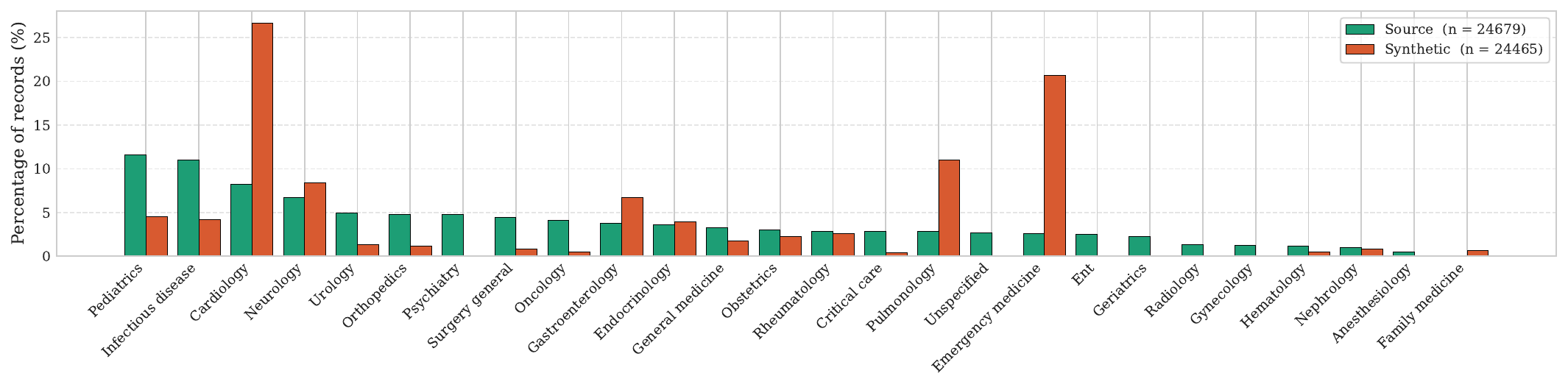}
    \caption{Specialty}
  \end{subfigure}\\[0.5em]
  \begin{subfigure}[t]{0.45\textwidth}
    \includegraphics[width=\linewidth]{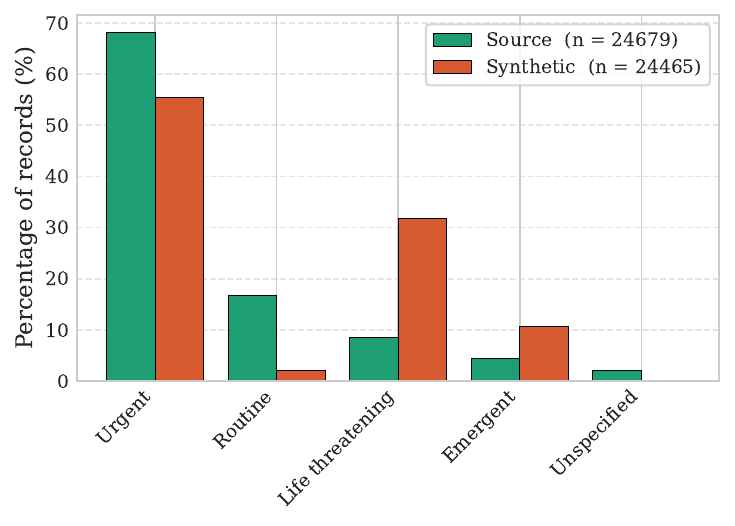}
    \caption{Urgency}
  \end{subfigure}\hfill
  \begin{subfigure}[t]{0.45\textwidth}
    \includegraphics[width=\linewidth]{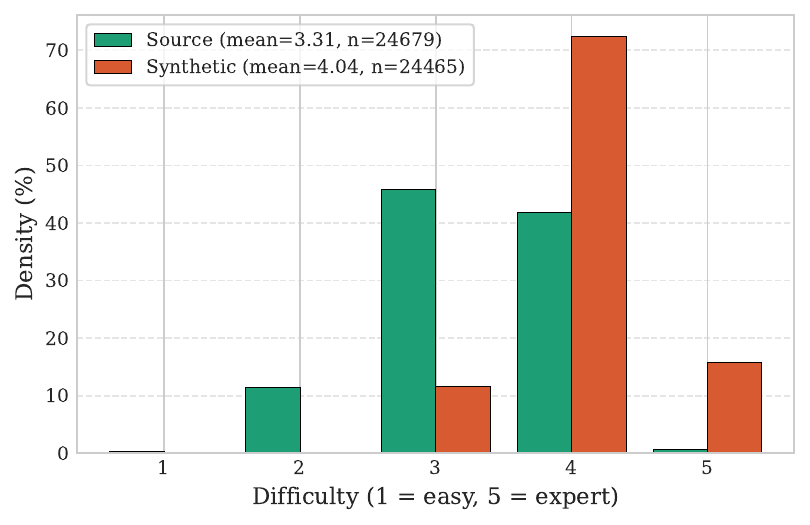}
    \caption{Difficulty}
  \end{subfigure}
  \caption{Synthetic MOOVE vs.\ source ($n_{\text{src}}=24{,}679$,
  $n_{\text{syn}}=24{,}465$). Top specialties preserved in rank;
  difficulty shifts toward levels $4$--$5$.}
  \label{fig:moove_dists}
\end{figure}

\begin{figure}[h]
  \centering
  \begin{subfigure}[t]{0.98\textwidth}
    \includegraphics[width=\linewidth]{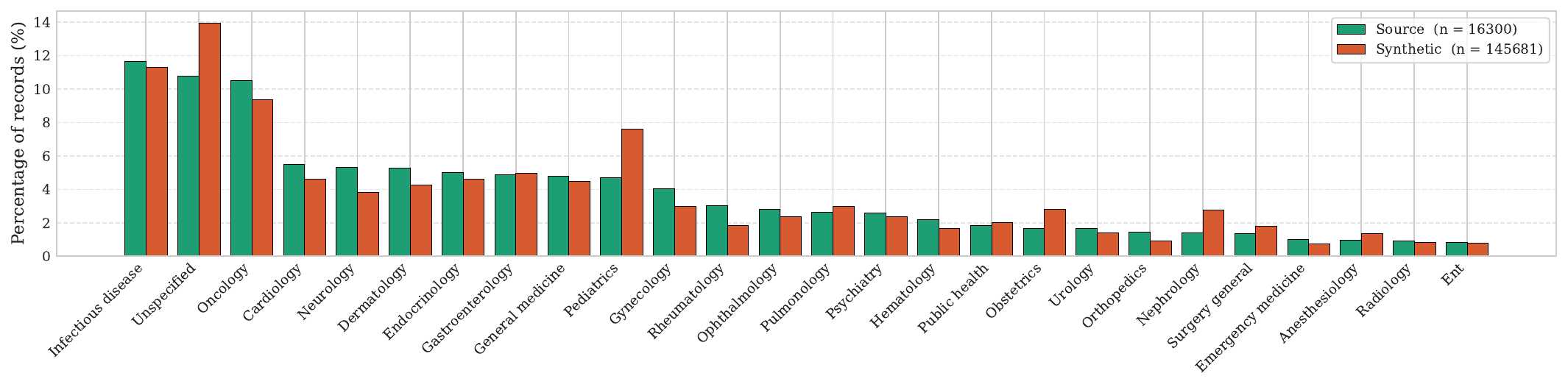}
    \caption{Specialty}
  \end{subfigure}\\[0.5em]
  \begin{subfigure}[t]{0.7\textwidth}
    \includegraphics[width=\linewidth]{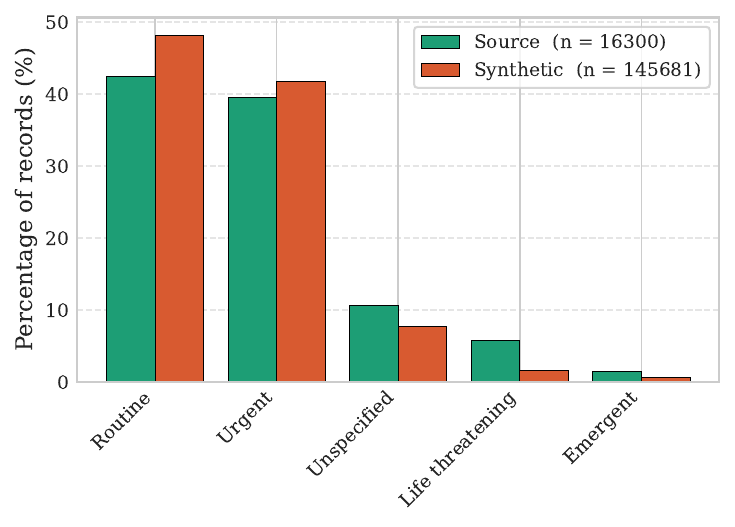}
    \caption{Urgency}
  \end{subfigure}
  \caption{Guidelines QA vs.\ source ($n_{\text{src}}=16{,}300$,
  $n_{\text{syn}}=145{,}681$, a $\sim\!9\times$ amplification). Difficulty
  is not comparable for this component, since the source consists of
  clinical practice guidelines rather than question--answer pairs. Both
  annotated axes closely match the source ($\mathrm{JSD}\le 0.014$).}
  \label{fig:guidelines_dists}
\end{figure}

\begin{figure}[H]
  \centering
  \begin{subfigure}[t]{0.98\textwidth}
    \includegraphics[width=\linewidth]{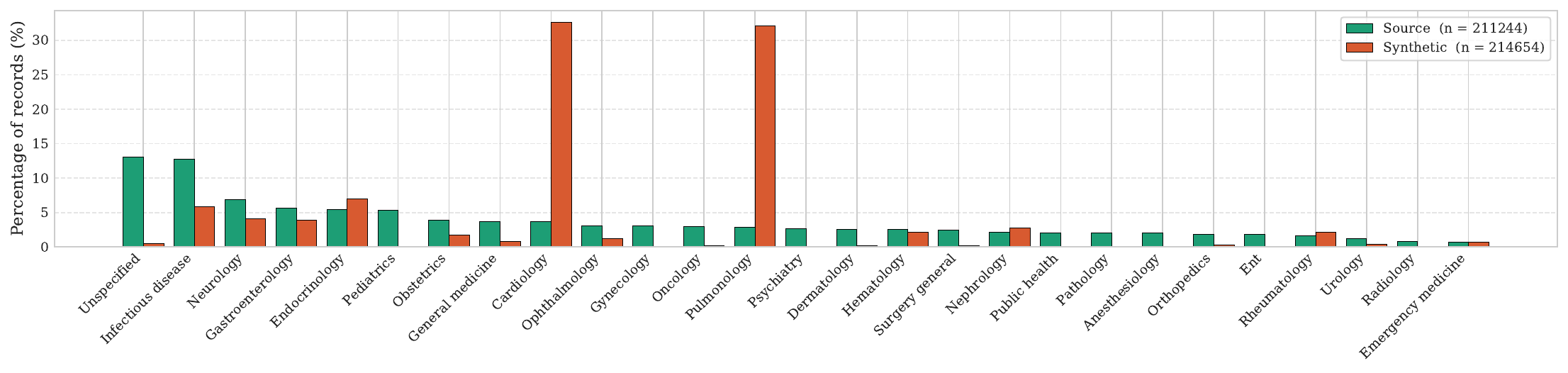}
    \caption{Specialty}
  \end{subfigure}\\[0.5em]
  \begin{subfigure}[t]{0.45\textwidth}
    \includegraphics[width=\linewidth]{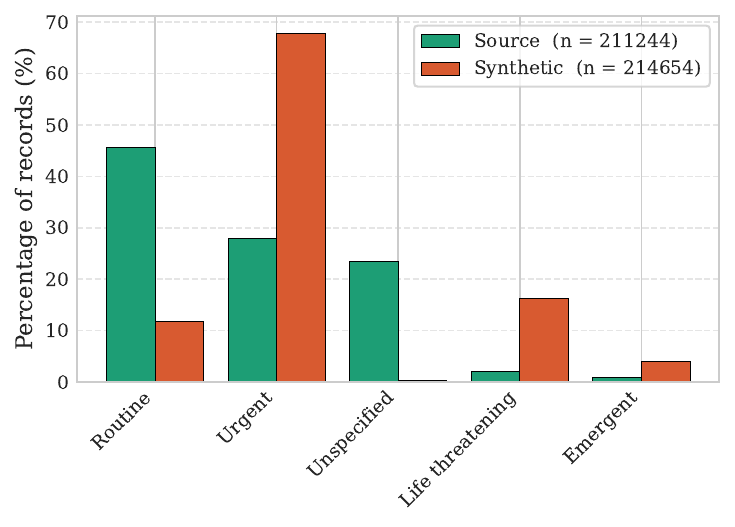}
    \caption{Urgency}
  \end{subfigure}\hfill
  \begin{subfigure}[t]{0.45\textwidth}
    \includegraphics[width=\linewidth]{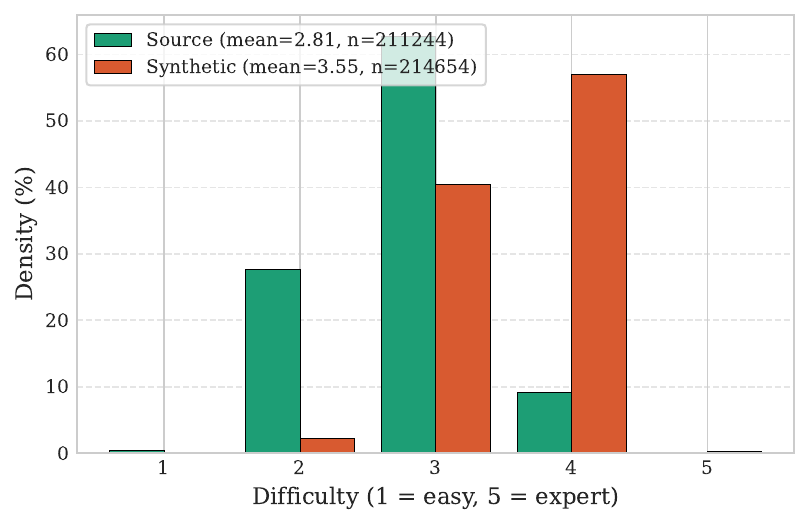}
    \caption{Difficulty}
  \end{subfigure}
  \caption{Synthetic Curated QA vs.\ source ($n_{\text{src}}=211{,}244$,
  $n_{\text{syn}}=214{,}654$). The generator broadens coverage from the
  eight aggregated source datasets, promoting under-represented
  specialties; difficulty shift is $2.81 \rightarrow 3.55$.}
  \label{fig:curated_dists}
\end{figure}

We compare each synthetic component against its source along three axes:
\textit{specialty}, \textit{urgency}, and \textit{difficulty}
(Figures~\ref{fig:moove_dists}--\ref{fig:curated_dists}). All labels are
produced by Qwen3-32B \citep{yang2025qwen3} as a zero-shot classifier with one
prompt template per axis; the same model labels source and synthetic, so
systematic classifier bias largely cancels in the comparison. Figure
captions report Jensen--Shannon divergence $\mathrm{JSD}$ for categorical
axes and the Wasserstein-1 distance $\mathcal{W}_1$ for difficulty (which
under a unimodal shift coincides with the mean shift). Components order
cleanly by source homogeneity: Guidelines (single corpus type) is the
tightest match, MOOVE (single vignette pool) is intermediate, and Curated
QA (eight aggregated public datasets) shows the largest redistribution.
Difficulty shifts upward by $\sim\!0.73$ points on the $1$--$5$ scale in
both MOOVE and Curated, indicating a near-uniform translation produced by
the generator rather than a dataset-specific effect.

% \begin{table}[h]
%   \centering
%   \small
%   \caption{Source vs.\ synthetic divergences. Lower is closer alignment.
%   Difficulty is not reported for Guidelines, whose source documents are
%   clinical practice guidelines rather than question--answer pairs.
%   Bracketed: $95\%$ bootstrap CIs.}
%   \label{tab:divergences}
%   \begin{tabular}{l l c c c}
%     \toprule
%     \textbf{Component} & \textbf{Axis} & \textbf{JSD} & \textbf{TVD} & $\mathcal{W}_1$ \\
%     \midrule
%     \multirow{3}{*}{Synthetic MOOVE}
%       & Specialty   & $0.260\,[0.254,\,0.267]$ & $0.502\,[0.494,\,0.510]$ & --- \\
%       & Urgency     & $0.122\,[0.118,\,0.127]$ & $0.294\,[0.288,\,0.302]$ & --- \\
%       & Difficulty  & ---                      & ---                      & $0.727\,[0.716,\,0.737]$ \\
%     \midrule
%     \multirow{2}{*}{Guidelines QA}
%       & Specialty   & $0.013\,[0.012,\,0.015]$ & $0.104\,[0.097,\,0.112]$ & --- \\
%       & Urgency     & $0.014\,[0.012,\,0.015]$ & $0.080\,[0.074,\,0.086]$ & --- \\
%     \midrule
%     \multirow{3}{*}{Synthetic Curated QA}
%       & Specialty   & $0.398\,[0.396,\,0.400]$ & $0.609\,[0.607,\,0.612]$ & --- \\
%       & Urgency     & $0.298\,[0.296,\,0.300]$ & $0.570\,[0.567,\,0.572]$ & --- \\
%       & Difficulty  & ---                      & ---                      & $0.747\,[0.744,\,0.750]$ \\
%     \bottomrule
%   \end{tabular}
% \end{table}

\newpage

\begin{table}[H]
\centering
\caption{Description of Curated QA}
\label{tab:training_datasets_curated}
\small
\resizebox{\textwidth}{!}{%
\begin{tabular}{l l r l}
\toprule
\textbf{Component} & \textbf{Dataset} & \textbf{No. examples} & \textbf{Description} \\
\midrule
\multirow{8}{*}{Curated QA}
 & MedQA          & 10{,}178  & USMLE-style exam questions (5-option MCQ) \\
 & MedMCQA        & 182{,}822 & Indian medical entrance exam questions \\
 & PubMedQA       & 1{,}000   & Biomedical QA from PubMed abstracts (balanced split) \\
 & MedExpQA       & 434       & Spanish medical residency exam (open-ended) \\
 & AfriMed-QA v1  & 3{,}000   & Pan-African multi-specialty (908 MCQ, 2{,}082 SAQ) \\
 & AfriMed-QA v2  & 15{,}275  & Pan-African (10k consumer, 4{,}039 MCQ, 1{,}236 SAQ) \\
 & HealthSearchQA & 3{,}375   & Common consumer medical queries from search engines \\
 & LiveQA         & 635       & Consumer health questions from the U.S. NLM \\
\midrule
\textbf{Total} & & \textbf{216{,}719} & \\
\bottomrule
\end{tabular}}
\end{table}

\section{Overview of Fully Open Meditron evaluation datasets.}
\label{appendix:evaluation_datasets}

\begin{table}[H]
\centering
\caption{Overview of Fully Open Meditron evaluation datasets.}
\label{tab:eval_datasets}
\resizebox{\textwidth}{!}{%
\begin{tabular}{l l r c @{\hspace{2em}} l r c}
\toprule
\multicolumn{4}{c}{\textbf{Medical evaluation}} & \multicolumn{3}{c}{\textbf{General purpose}} \\
\cmidrule(lr){1-4} \cmidrule(lr){5-7}
\textbf{Task} & \textbf{Dataset} & \textbf{No. examples} & \textbf{OOD} & \textbf{Dataset} & \textbf{No. examples} & \textbf{OOD} \\
\midrule
\multirow{4}{*}{Medical QA (MCQ)}
 & MedQA       & 1{,}273 & --          & MMLU-Pro      & 12{,}032 & N/A \\
 & MedMCQA     & 4{,}183 & --          & IFEval        & 541      & N/A \\
 & PubMedQA    & 500     & --          & ARC-Challenge & 1{,}172  & N/A \\
 & MedXpertQA  & 2{,}450 & \checkmark  &               &          &     \\
\cmidrule(lr){1-4}
Open-ended clinical & Auto-MOOVE & 24{,}681 & \checkmark & & & \\
\cmidrule(lr){1-4}
Open-ended rubric   & HealthBench Hard & 1{,}000 & \checkmark & & & \\
\bottomrule
\end{tabular}%
}
\end{table}

\section{Medical benchmark accuracy}
\label{appendix:medical_benchmark_accuracy}

\begin{table}[H]
\centering
\caption{Medical benchmark accuracy (\%). Judge is Qwen3-30B-A3B-Instruct. Best within partition \textbf{bolded}; best fully open \underline{underlined}. }
\resizebox{\textwidth}{!}{%
\begin{tabular}{lccccccc}
\toprule
\textbf{Model} & \textbf{MedMCQA} & \textbf{MedQA} & \textbf{PubMedQA} & \textbf{MedXpertQA} & \textbf{Healthbench} & \textbf{Avg} & \textbf{Gain} \\
\midrule
\multicolumn{8}{l}{\textit{Closed/Open-access reference}} \\
gpt-oss-120b               & 68.85 & 88.30 & 75.00 & 31.76 & 48.60 & \textbf{62.50}  & --    \\
Qwen3-30B-A3B-Instruct-2507 & 70.95 & 85.23 & 74.80 & 23.06 & 43.02 & 59.41 &  --   \\
Llama-3.1-70B              & 71.67 & 84.29 & 77.4 & 23.71 & 20.30 & 55.47 & --    \\
Llama-3.1-70B-Meditron             & 72.17 & 84.6 & 77.8 & 26.82 & 13.05 & 54.89 & -0.58  \\
Phi-3.5-mini-instruct      & 53.14 & 57.19 & 72.00 & 13.22 & 28.84 & 44.88 & --    \\
MediPhi                    & 53.65 & 57.5 & 72.8 & 12.29 & 25.29 & 44.31 & -0.57 \\
\midrule
\multicolumn{8}{l}{\textit{Open-access base $\rightarrow$ Fully Open Meditron finetune}} \\
gemma-3-27b-it             & 62.75 & 76.20 & 74.60 & 16.69 & 45.78 & 55.20 & --   \\
\quad + Fully Open Meditron      & 63.71 & 77.61 & 75.80 & 18.00 & 47.15 & 56.45 & 1.25 \\
MedGemma-27B               & 66.44 & 86.10 & 73.00 & 21.88 & 41.95 & \textbf{57.87} & 2.67 \\
\midrule
\multicolumn{8}{l}{\textit{Fully open base $\rightarrow$ Fully Open Meditron finetune}} \\
Apertus-70B-Instruct       & 52.43 & 60.64 & 66.80 & 12.33 & 32.28 & 44.90 & --    \\
\quad + \underline{Fully Open Meditron}      & 56.32 & 68.58 & 75.20 & 16.90 & 40.14 & \underline{\textbf{51.43}} & 6.53  \\
%\quad + Fully Open Meditron (with replay) & 57.11 & 71.72 & 72.6 & 15.22 & 32.58 & 49.85 & 4.82  \\
Olmo-2-32B-SFT             & 59.10 & 66.22 & 72.00 & 13.02 & 19.75 & 45.88 & --    \\
\quad + Fully Open Meditron & 57.83 & 69.44 & 76.60 & 17.96 & 33.82 & 51.13 & 5.26  \\
EuroLLM-22B-Instruct       & 54.94 & 66.61 & 73.60 & 14.61 & 34.79 & 48.91 & --    \\
\quad + Fully Open Meditron      & 54.79 & 63.16 & 78.00 & 14.61 & 37.38 & 49.59 & 0.68  \\
\midrule
\multicolumn{8}{l}{\textit{Small Fully open base $\rightarrow$ Fully Open Meditron finetune}} \\
Apertus-8B-Instruct        & 45.80 & 51.14 & 37.60 & 11.71 & 21.55 & 33.56 & --    \\
\quad + Fully Open Meditron      & 48.74 & 58.44 & 75.60 & 13.67 & 38.11 & \textbf{46.91} & 13.35 \\
EuroLLM-9B-Instruct        & 37.84 & 48.55 & 40.00 & 10.33 & 13.47 & 30.04 & --    \\
\quad + Fully Open Meditron      & 46.98 & 49.73 & 67.40 & 11.63 & 31.62 & 41.47 & 11.43 \\
\bottomrule
\end{tabular}}
\footnotesize{\textit{Note:} Per-task 95\% CIs (approximate, varying with $p$): MedMCQA $\pm$1.5 pp ($n{=}4183$), MedQA $\pm$2.6 pp ($n{=}1273$), PubMedQA $\pm$3.9 pp ($n{=}500$), MedXpertQA $\pm$1.5 pp ($n{=}2450$), Healthbench Hard $\pm$2.9 pp ($n{=}1000$). Avg CIs computed by SE propagation. Gains are paired differences; the unpaired SE bound is $\sim$1.2 pp, so gains $>$2.5 pp are robustly significant.}
\end{table}

\section{Auto-MOOVE pairwise results }
\label{appendix:auto_moove_results}

\begin{table}[H]
\centering
\caption{Auto-MOOVE pairwise results ($N=12{,}602$ per pair). Net Win Rate $=$ Win$-$Loss \%; Adjusted Win Rate $=$ Win $+$ Tie/2; $\Delta$Likert averaged across criteria.}
\resizebox{\textwidth}{!}{%
\begin{tabular}{llccc}
\toprule
\textbf{Base / Comparison} & \textbf{Our model} & \textbf{Net Win Rate} & \textbf{Adj. Win Rate} & \textbf{$\Delta$Likert} \\
\midrule
\multicolumn{5}{l}{\textit{Corresponding base vs.\ *-Meditron }} \\
Apertus-70B-Instruct    & Apertus-70B-MeditronFO       & +59.3 & 79.6 & +0.40 \\
%Apertus-70B-Instruct    & Apertus-70B-MeditronFO (w. replay)  & +15.3 & 57.7 & +0.02 \\
OLMo-2-32B-SFT          & OLMo-2-32B-MeditronFO        & +67.2 & 83.7 & +0.43 \\
EuroLLM-22B-Instruct    & EuroLLM-22B-MeditronFO       & \phantom{0}+8.0 & 54.0 & +0.04 \\
\midrule
\multicolumn{5}{l}{\textit{Small models — base vs. *-Meditron}} \\
EuroLLM-9B-Instruct     & EuroLLM-9B-MeditronFO        & +72.5 & 86.3 & +0.66 \\
Apertus-8B-Instruct     & Apertus-8B-Meditron        & +75.7 & 87.8 & +0.62 \\
\midrule
\multicolumn{5}{l}{\textit{Comparison with Meditron 3}} \\
Llama-3.1-70B-Meditron          & Apertus-70B-MeditronFO       & +93.2 & 96.6 & +0.94 \\
%Llama-3.1-70B-Meditron          & Apertus-70B-MeditronFO (w. replay)  & +76.8 & 88.4 & +0.71 \\
\midrule
\multicolumn{5}{l}{\textit{Comparison with MedGemma}} \\
Gemma-3-27B-IT          & Gemma-3-27B-MeditronFO       & +29.8 & 64.9 & +0.15 \\
MedGemma                & Gemma-3-27B-MeditronFO       & +32.7 & 66.3 & +0.16 \\
MedGemma          & Gemma-3-27B-IT (not ours)                 & \phantom{0}+5.8 & 52.9 & +0.02 \\
MedGemma                & Apertus-70B-MeditronFO       & +24.7 & 62.4 & +0.11 \\
\bottomrule
\end{tabular}%
}
\end{table}

\begin{table}[h]
\centering
\caption{Auto-MOOVE pairwise comparisons ($N=12{,}602$ per pair), Judge ablations.}
\label{tab:judge1}
\resizebox{\textwidth}{!}{%
\begin{tabular}{lllccc}
\toprule
\textbf{Base model} & \textbf{Our model} & \textbf{Judge} & \textbf{Net Win Rate} & \textbf{Adj. Win Rate} & \textbf{$\Delta$Likert} \\
\midrule
OLMo-2-32B-SFT          & OLMo-2-32B-MeditronFO   & Qwen3-30B-A3B      & +67.2 & 83.7 & +0.43 \\
OLMo-2-32B-SFT          & OLMo-2-32B-MeditronFO   & Qwen3-235B-A22B    & +69.7 & 84.8 & +0.44 \\
OLMo-2-32B-SFT          & OLMo-2-32B-MeditronFO   & gpt-oss-120        & +35.1 & 67.5 & +0.32 \\
OLMo-2-32B-SFT           & OLMo-2-32B-MeditronFO - Synth. MOOVE& Qwen3-30B-A3B   & +56.2 & 78.2 & +0.31 \\
\midrule
EuroLLM-22B-Instruct    & EuroLLM-22B-MeditronFO  & Qwen3-30B-A3B      & \phantom{0}+8.0 & 54.0 & +0.04 \\
EuroLLM-22B-Instruct    & EuroLLM-22B-MeditronFO  & Qwen3-235B-A22B    & +34.3 & 67.2 & +0.20 \\
EuroLLM-22B-Instruct    & EuroLLM-22B-MeditronFO  & gpt-oss-120        & -12.5 & 43.7 & -0.15 \\
EuroLLM-22B-Instruct     & EuroLLM-22B-MeditronFO - Synth. MOOVE& Qwen3-30B-A3B   & \phantom{0}+7.3 & 53.7 & +0.02 \\
\midrule
Gemma-3-27B-IT          & Gemma-3-27B-MeditronFO  & Qwen3-30B-A3B      & +29.8 & 64.9 & +0.15 \\
MedGemma                & Gemma-3-27B-MeditronFO & Qwen3-30B-A3B       & +32.7 & 66.3 & +0.16 \\
MedGemma          & Gemma-3-27B-MeditronFO  & Qwen3-235B-A22B    & +17.2 & 58.6 & +0.04 \\
Gemma-3-27B-IT          & Gemma-3-27B-MeditronFO  & gpt-oss-120        & +23.0 & 61.5 & +0.11 \\
Gemma-3-27B-IT           & Gemma-3-27B-MeditronFO - Synth. MOOVE& Qwen3-30B-A3B   & +25.0 & 62.5 & +0.12 \\
\bottomrule
\end{tabular}}
\end{table}

\section{General-purpose benchmark results. }
\label{appendix:generalist}

\begin{table}[H]
\centering
\caption{General-purpose benchmark results. }
\begin{tabular}{lcccc}
\toprule
\textbf{Model} & \textbf{MMLU-Pro} & \textbf{IFEval} & \textbf{ARC Challenge} & \textbf{Avg} \\
\midrule
\multicolumn{5}{l}{\textit{Closed/Open-access reference}} \\
Llama-3.1-70B              & 65.91 & 84.66 & 62.54 & 71.04 \\
Llama-3.1-70B-Meditron             & 24.38 & 58.04 & 54.35 & 45.59 \\
Phi-3.5-mini-instruct      & 46.58 & 49.54 & 50.94 & 49.02 \\
MediPhi                    & 48.45 & 37.89 & 50.51 & 45.62 \\
gemma-3-27b-it             & 65.17 & 76.34 & 59.22 & 66.91 \\
\quad + Fully Open Meditron      & 62.17 & 53.42 & 62.46 & 59.35 \\
MedGemma-27B               & 62.51 & 72.46 & 57.42 & 64.13 \\
\midrule
\multicolumn{5}{l}{\textit{Fully open base $\rightarrow$ Fully Open Meditron finetune}} \\
Apertus-70B-Instruct       & 43.25 & 64.70 & 54.35 & 54.10 \\
\quad + Fully Open Meditron      & 32.80 & 41.04 & 48.38 & 40.74 \\
%\quad + Fully Open Meditron (w. replay) & 37.13 & 61.92 & 51.28 & 50.11 \\
Olmo-2-32B-SFT             & 41.71 & 68.58 & 55.29 & 55.19 \\
\quad + Fully Open Meditron      & 45.55 & 68.21 & 56.91 & 56.89 \\
EuroLLM-22B-Instruct       & 34.95 & 71.72 & 51.19 & 52.62 \\
\quad + Fully Open Meditron      & 38.51 & 61.55 & 54.61 & 51.56 \\
\midrule
\multicolumn{5}{l}{\textit{Small Fully open base $\rightarrow$ Fully Open Meditron finetune}} \\
Apertus-8B-Instruct        & 34.33 & 68.58 & 51.71 & 51.54 \\
\quad + Fully Open Meditron      & 28.74 & 58.41 & 48.46 & 45.20 \\
EuroLLM-9B-Instruct        & 27.41 & 59.52 & 50.09 & 45.67 \\
\quad + Fully Open Meditron      & 25.58 & 50.65 & 48.29 & 41.51 \\
\bottomrule
\end{tabular}
\end{table}

\section{Additional Ablations }
\label{appendix:ablation_study}

\textbf{General-purpose capability as a smoke test.}
We treat general-purpose benchmarks as a smoke test for catastrophic forgetting rather than as a primary optimization target. Domain adaptation is expected to trade off some broad instruction-following capability against improved medical specialization~\cite{xie2025medical}, and our results should be interpreted in that light. By default, Apertus-70B-MeditronFO drops 13.4 points on the general-purpose average relative to its base (54.10 $\rightarrow$ 40.74), driven largely by IFEval (64.70 $\rightarrow$ 41.04). However, this pattern is neither unique to our recipe nor uniformly severe across models: OLMo-2-32B-MeditronFO improves slightly over its base (+1.70), while smaller models and Gemma-3-27B show moderate degradations. Notably, MedGemma-27B also underperforms its general-purpose base Gemma-3-27B, indicating that this tradeoff is a broader feature of medical specialization rather than a pathology of fully open training. Relative to prior open medical finetunes, our recipe also appears to retain more general capability: Llama-3.1-70B-Meditron exhibits a substantially larger drop than Apertus-70B-MeditronFO (45.59 vs.\ 71.04 for its base), suggesting that the cost of specialization is reduced, though not eliminated, in our setting. Detailed results are recorded in Table \ref{tab:generalist} 

As an optional mitigation, the training mixture can be augmented with a 10\% subset of the fully open Tülu 3 SFT mixture~\cite{lambert2024tulu}, which recovers most of the general-purpose loss for Apertus-70B (49.85 average, 61.92 on IFEval) while largely preserving medical gains. We do not include Tülu replay in the default Fully Open Meditron recipe, because our primary objective is domain specialization and we prefer to keep the core corpus focused and interpretable. Instead, we provide instructions for enabling replay in the codebase and document the corresponding ablation.

\begin{table}[h]
\centering
\caption{General-purpose benchmark results. For a detailed table with older closed open access reference please see appendix \ref{appendix:generalist}}
\label{tab:generalist}
\begin{tabular}{lcccc}
\toprule
\textbf{Model} & \textbf{MMLU-Pro} & \textbf{IFEval} & \textbf{ARC Challenge} & \textbf{Avg} \\
\midrule
\multicolumn{5}{l}{\textit{Fully open base $\rightarrow$ Fully Open Meditron finetune}} \\
Apertus-70B-Instruct       & 43.25 & 64.70 & 54.35 & 54.10 \\
\quad + Fully Open Meditron      & 32.80 & 41.04 & 48.38 & 40.74 \\
%\quad + Fully Open Meditron (w. replay) & 37.13 & 61.92 & 51.28 & 50.11 \\
Olmo-2-32B-SFT             & 41.71 & 68.58 & 55.29 & 55.19 \\
\quad + Fully Open Meditron      & 45.55 & 68.21 & 56.91 & 56.89 \\
EuroLLM-22B-Instruct       & 34.95 & 71.72 & 51.19 & 52.62 \\
\quad + Fully Open Meditron      & 38.51 & 61.55 & 54.61 & 51.56 \\
\midrule
\multicolumn{5}{l}{\textit{Small Fully open base $\rightarrow$ Fully Open Meditron finetune}} \\
Apertus-8B-Instruct        & 34.33 & 68.58 & 51.71 & 51.54 \\
\quad + Fully Open Meditron      & 28.74 & 58.41 & 48.46 & 45.20 \\
EuroLLM-9B-Instruct        & 27.41 & 59.52 & 50.09 & 45.67 \\
\quad + Fully Open Meditron      & 25.58 & 50.65 & 48.29 & 41.51 \\
\bottomrule
\end{tabular}
\end{table}

%\textbf{Ablation of Synthetic Moove for Open ended evaluation. } \xavier{todo}

%Meditron3-70B            & Apertus-70B-MeditronFO  & +92.2 & 96.1 & +0.92  \\
%MedGemma-27B-IT          & Gemma-3-27B-MeditronFO  & +28.6 & 64.3 & +0.13 \\
%MedGemma-27B-IT          & + Fully Open Meditron   & +28.6 & 64.3 & +0.13 \\

\begin{table}[h]
\centering
\caption{\textbf{Ablation of Judge for Open ended evaluation.} Across eight diverse judges spanning model families (Qwen, GPT-OSS, Gemma, GLM, Llama, Nemotron) and sizes (27B--235B), Apertus-70B-MeditronFO is consistently preferred over Apertus-70B-Instruct, with adjusted win rates ranging from 73.2\% (Llama-3.3-70B) to 93.7\% (Nemotron-3-Nano-30B) and all Likert deltas strictly positive. Notably, gpt-oss-120b is the model used for our synthetic data generation; using the generator as a judge would conceptually favor models stylistically closer to its own outputs, yet it does not yield disproportionately higher win rates than other independent judges. This argues against a style-matching explanation for the observed gains. Auto-MOOVE pairwise comparisons ($N=12{,}602$ per pair); complementary results are in appendix~\ref{appendix:auto_moove_results}.}
\label{tab:judge}
\resizebox{\textwidth}{!}{%
\begin{tabular}{lllccc}
\toprule
\textbf{Base model} & \textbf{Our model} & \textbf{Judge} & \textbf{Net Win Rate} & \textbf{Adj. Win Rate} & \textbf{$\Delta$Likert} \\
\midrule
Apertus-70B-Instruct    & Apertus-70B-MeditronFO & Qwen3-30B-A3B      & +59.3 & 79.6 & +0.40 \\
Apertus-70B-Instruct    & Apertus-70B-MeditronFO & Qwen3-235B-A22B    & +75.4 & 87.7 & +0.52 \\
Apertus-70B-Instruct    & Apertus-70B-MeditronFO & gpt-oss-120        & +52.2 & 76.1 & +0.53 \\
Apertus-70B-Instruct & Apertus-70B-MeditronFO & Gemma-3-27B-it         & +80.9 & 90.5 & +0.59 \\
Apertus-70B-Instruct & Apertus-70B-MeditronFO & GLM-4.5-Air            & +85.1 & 92.5 & +0.93 \\
Apertus-70B-Instruct & Apertus-70B-MeditronFO & Llama-3.3-70B-Instruct & +46.3 & 73.2 & +0.52 \\
Apertus-70B-Instruct & Apertus-70B-MeditronFO & Nemotron-3-Super-120B  & +71.9 & 86.0 & +0.83 \\
Apertus-70B-Instruct & Apertus-70B-MeditronFO & Nemotron-3-Nano-30B    & +87.5 & 93.7 & +0.90 \\
\bottomrule
\end{tabular}}
\end{table}

\begin{table}[H]
\centering
\caption{Ablation study. All runs use Apertus-70B-Instruct as base. Judge is Qwen3-30B-A3B-Instruct}
\resizebox{\textwidth}{!}{%
\begin{tabular}{lcccccccccccc}
\toprule
\textbf{Configuration} & \textbf{MedMCQA} & \textbf{MedQA} & \textbf{PubMedQA} & \textbf{HealthBench} & \textbf{MedXpertQA} & \textbf{Med Avg} & \textbf{MMLU-Pro} & \textbf{IFEval} & \textbf{ARC-C} & \textbf{Auto-MOOVE} & \textbf{$\Delta$ Likert} \\
\midrule
Apertus-70B-Instruct & 52.43 & 60.64 & 66.80 & 32.28 & 12.33 & 44.90 & 43.25 & 64.70 & 54.35 & -- & -- \\
\midrule
\multicolumn{12}{l}{\textit{Corpus component ablations (one component removed at a time)}} \\
Full corpus (default) & 56.32 & 68.58 & 75.20 & 40.14 & 16.90 & 51.43 & 32.80 & 41.04 & 48.38 & 79.6 & 0.40 \\
$-$ Curated QA & 54.36 & 65.12 & 71.8 & 30.34 & 17.18 & 47.76 & 39.3 & 39.56 & 48.72 & 73.4 & 0.27 \\
$-$ Synthetic Curated QA & 55.03 & 66.46 & 74.4 & 41.32 & 15.59 & 50.56 & 28.73 & 40.85 & 49.06 & 78.7 & 0.4 \\
$-$ Guidelines QA & 56.35 & 69.6 & 76.2 & 42.35 & 17.35 & 52.37 & 12.77 & 36.41 & 45.9  & 78.7 & 0.39 \\
$-$ Synthetic MOOVE & 56.37 & 69.91 & 72.60 & 39.85 & 18.20 & 54.27 & 29.56 & 33.46 & 47.87 & 75.5 & 0.34 \\
\midrule
\multicolumn{12}{l}{\textit{Replay data ablation (Tülu fraction)}} \\
0\% (default) & 56.32 & 68.58 & 75.20 & 40.14 & 16.90 & 51.43 & 32.80 & 41.04 & 48.38 & 79.6 & 0.40 \\
10\% Tülu & 57.11 & 71.72 & 72.60 & 32.58 & 15.22 & 49.85 & 37.13 & 61.92 & 51.28 & 57.7 & 0.02 \\
\midrule
\multicolumn{12}{l}{\textit{Teacher choice for distillation}} \\
gpt-oss-120b (default) & 56.32 & 68.58 & 75.20 & 40.14 & 16.90 & 51.43 & 32.80 & 41.04 & 48.38 & 79.6 & 0.40 \\
qwen-3-30b & 60.67 & 73.13 & 73.80 & 14.40 & 17.27 & 56.22 & 40.04 & 57.86 & 50.00 & 41.7 & -0.48 \\
\bottomrule
\end{tabular}}
\end{table}

\section{AutoMOOVE validation}
\label{appendix:automoove_validation}

\begin{table}[H]
\centering
\caption{Auto-MOOVE validation against human clinical judgments. \textit{Left:} judge $\kappa$ against the full human panel, situated within the distribution of per-rater $\kappa$ values (each rater scored against the consensus of all others; minimum 10 triplets per rater). \textit{Right:} average Likert score difference (chosen minus rejected) per criterion.}
\label{tab:auto_moove}
\begin{minipage}[t]{0.48\textwidth}
    \centering
    \resizebox{\textwidth}{!}{%
    \begin{tabular}{lcc}
    \toprule
    & \textbf{With Ties} & \textbf{No Ties} \\
    \midrule
    Human raters ($n$)         & 204                & 156                \\
    Human $\kappa$ mean (std)  & 0.320 (0.228)      & 0.544 (0.285)      \\
    Human $\kappa$ median      & 0.292              & 0.526              \\
    Human $\kappa$ 95\% CI     & [0.289, 0.353]     & [0.497, 0.586]     \\
    \midrule
    Judge $\kappa$ (95\% CI)   & 0.232 {\scriptsize[0.209--0.254]} & 0.487 {\scriptsize[0.447--0.526]} \\
    Judge percentile           & 39.7\%             & 42.9\%             \\
    Judge z-score              & $-0.39\sigma$      & $-0.20\sigma$      \\
    \bottomrule
    \end{tabular}}
\end{minipage}\hfill
\begin{minipage}[t]{0.48\textwidth}
    \centering
    \resizebox{\textwidth}{!}{%
    \begin{tabular}{lcc}
    \toprule
    \textbf{Criterion} & \textbf{Judge} & \textbf{Human} \\
    \midrule
    Question comprehension    & +0.23 & +0.68 \\
    Logical reasoning         & +0.30 & +0.77 \\
    Relevance \& completeness & +0.32 & +1.05 \\
    Harmlessness              & +0.03 & +0.76 \\
    Fairness                  & +0.03 & +0.48 \\
    Contextual awareness      & +0.28 & +0.74 \\
    Communication             & +0.24 & +0.66 \\
    Clarity                   & +0.22 & +0.87 \\
    Align. with guidelines    & +0.28 & +0.84 \\
    \bottomrule
    \end{tabular}}
\end{minipage}
\end{table}

\begin{figure}[H]
    \centering
    \includegraphics[width=0.6\textwidth]{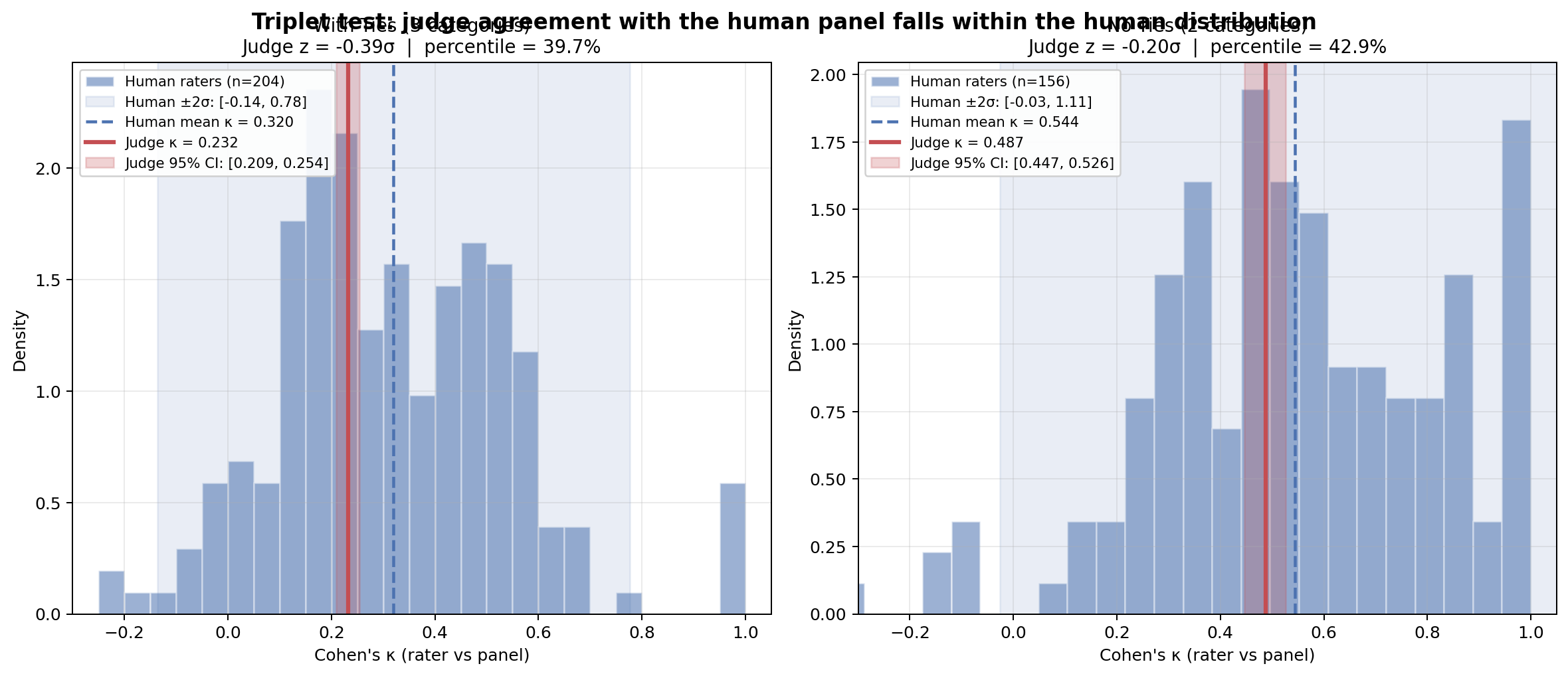}
    \caption{Distribution of per-rater $\kappa$ values across the 204-rater human panel, with the Auto-MOOVE judge's $\kappa$ situated within it. The judge falls within $\pm 2\sigma$ of the human mean under both with-ties and no-ties scoring, indicating it is statistically indistinguishable from a typical human rater on this validation set.}
    \label{fig:judge_indistinguishable}
\end{figure}

\section{Training details}
\label{appendix:training}

\subsection{Infrastructure and framework}

All Fully Open Meditron models were trained on a high-performance computing cluster using NVIDIA GH200 Grace Hopper Superchip nodes with 4 GPUs per node. Large bases (Apertus-70B, OLMo-2-32B, EuroLLM-22B, Gemma-3-27B) were trained on 8 nodes (32 GH200 GPUs); small bases (Apertus-8B, EuroLLM-9B) were trained on 4 nodes (16 GH200 GPUs).

Training used the Axolotl framework with PyTorch's \texttt{torchrun} launcher and \texttt{c10d} rendezvous. The 70B Apertus run used DeepSpeed ZeRO Stage 3 for memory partitioning; all other runs used PyTorch FSDP v2 with transformer-block auto-wrap, sharded state-dict checkpointing, reshard-after-forward, and activation checkpointing. Apertus-70B, OLMo-2-32B, and EuroLLM-22B/9B additionally used the cut-cross-entropy plugin to reduce activation memory at the loss-computation step. All runs used Flash Attention~2 and bfloat16 mixed-precision training.

\subsection{Common training settings}

To preserve the alignment work invested in each base, we maintained the instruction-tuning chat template native to each model (ChatML for EuroLLM via explicit override; native templates for all others). All runs share the following settings unless noted otherwise:

\begin{itemize}
    \item \textbf{Sequence length:} 4096 tokens with sample packing.
    \item \textbf{Optimizer:} AdamW (fused implementation), $\beta_1 = 0.9$, $\beta_2 = 0.999$ (default) unless otherwise stated.
    \item \textbf{LR scheduler:} cosine decay with warmup.
    \item \textbf{Gradient clipping:} max gradient norm 1.0.
    \item \textbf{Random seeds:} 42 for both model initialization and data shuffling.
\end{itemize}

\subsection{Per-model hyperparameters}

Per-model settings are summarized in Table~\ref{tab:training_hparams}. Hyperparameters were selected based on each base model's published instruction-tuning recipe where available, and lightly tuned via short pilot runs on a held-out subset of MedQA/MedMCQA dev splits before full training.

\begin{table}[h]
\centering
\caption{Per-model training hyperparameters. ``Eff.\ batch'' is the effective batch size in sequences (micro-batch $\times$ gradient accumulation $\times$ world size). All runs use fused AdamW, cosine LR schedule, sequence length 4096 with sample packing, and seed 42.}
\label{tab:training_hparams}
\small
\setlength{\tabcolsep}{4pt}
\begin{tabular}{lccccccccc}
\toprule
Model & Nodes (GPUs) & Parallelism & $\mu$BS & Accum. & Eff.\ batch & LR & Warmup & WD & $\beta_2$ \\
\midrule
Apertus-70B-Instruct  & 8 (32) & ZeRO-3  & 4 & 1 & 128 & 1e-5 & 0.10 & 0.0  & 0.999 \\
OLMo-2-32B-SFT        & 8 (32) & FSDP v2 & 1 & 8 & 256 & 5e-6 & 0.03 & 0.0  & 0.999 \\
EuroLLM-22B-Instruct  & 8 (32) & FSDP v2 & 1 & 8 & 256 & 1e-5 & 0.10 & 0.01 & 0.999 \\
Gemma-3-27B-IT        & 8 (32) & FSDP v2 & 1 & 8 & 256 & 1e-6 & 0.10 & 0.10 & 0.95  \\
Apertus-8B-Instruct   & 4 (16) & FSDP v2 & 8 & 1 & 128 & 2e-5 & 0.10 & 0.10 & 0.99  \\
EuroLLM-9B-Instruct   & 4 (16) & FSDP v2 & 4 & 2 & 128 & 1e-5 & 0.10 & 0.01 & 0.999 \\
\bottomrule
\end{tabular}
\end{table}

\subsection{Reproducibility artifacts}

The full Axolotl YAML configuration files for each model, the SLURM submission script, and the data preparation pipeline will be released alongside the corpus upon publication. A 10\% T\"ulu 3 SFT replay variant is also provided as an opt-in configuration but is not part of the default Fully Open Meditron recipe.

\subsection{Compute resources}
\label{app:compute}

\begin{table}[h]
\centering
\small
\caption{Compute for the main MeditronFO training runs. Wall-clock times are taken from cluster job logs. GPU-hours = nodes $\times$ 4 $\times$  wall-clock hours.}
\label{tab:compute_main}
\begin{tabular}{lccc}
\toprule
Model & Nodes (GPUs) & Wall-clock (h:mm) & GPU-hours \\
\midrule
Apertus-70B-MeditronFO     & 8 (32) & 6:39 & 213 \\
OLMo-2-32B-MeditronFO      & 8 (32) & 5:34 & 178 \\
EuroLLM-22B-MeditronFO     & 8 (32) & 3:45 & 120 \\
Gemma-3-27B-MeditronFO     & 8 (32) & 5:02 & 161 \\
Apertus-8B-MeditronFO      & 4 (16) & 2:30 &  40 \\
EuroLLM-9B-MeditronFO      & 4 (16) & 0:43 &  11 \\
\midrule
Total (main runs)          &        &      & 723 \\
\bottomrule
\end{tabular}
\end{table}

\section{Synthetic Data Generation Prompts}
\label{appendix:review_details}

This appendix documents the exact prompts used by the three synthetic data generation pipelines (Synthetic Curated QA, Guidelines QA, and Synthetic MOOVE). All three pipelines share a common system message and a component-specific developer message, followed by a user message that injects either few-shot exemplars or a source guideline. Prompts are reproduced verbatim from the generation scripts; placeholders such as \texttt{\{date\}}, \texttt{\{reasoning\}}, and the example slots are filled at runtime.

\subsection{Shared System Message}

All three pipelines use the harmony-format system message below, with \texttt{\{date\}} set to the generation date and \texttt{\{reasoning\}} set to \texttt{low}.

\begin{tcolorbox}[breakable, colback=gray!5, colframe=gray!50, title=System Message (shared)]
\begin{verbatim}
You are ChatGPT, a large language model trained by OpenAI.
Knowledge cutoff: 2024-06
Current date: {date}
Reasoning: {reasoning}

# Valid channels: analysis, commentary, final.
# Channel must be included for every message.
\end{verbatim}
\end{tcolorbox}

\subsection{Guidelines QA Prompt}
\label{appendix:prompt_guidelines}

The Guidelines QA pipeline seeds generation with one full clinical practice guideline per call and elicits ten multiple-choice vignettes grounded strictly in that document.

\begin{tcolorbox}[breakable, colback=gray!5, colframe=gray!50, title=Developer Message --- Guidelines QA]
\begin{verbatim}
You are an expert medical educator and physician tasked with creating
high-quality, clinically accurate content for a medical exam. Your task
is to generate clinical vignette-style questions along with its correct
answer, based STRICTLY on the provided medical guideline. Focus on
realistic patient presentations (age, symptoms, physical exam findings),
identifying 'red flags', and diagnostic reasoning highlighted in the
text. The timeline and objective progress should always be clear and
detailed in the vignettes. Include clear context about site and where
people travelled etc. Do not include outside information or unproven
treatments.
\end{verbatim}
\end{tcolorbox}

\begin{tcolorbox}[breakable, colback=gray!5, colframe=gray!50, title=User Message --- Guidelines QA]
\begin{verbatim}
Here is the medical guideline:

=== GUIDELINE START ===
{guideline_text}
=== GUIDELINE END ===

Based ONLY on the guideline above, generate exactly 10 unique
MULTIPLE-CHOICE clinical vignette questions and their answers. Each
question should present a realistic patient scenario that tests the
diagnostic or management principles in the text. For each vignette,
provide 4-5 plausible multiple-choice options (A-E). Ensure distractors
represent common diagnostic pitfalls or 'next best steps' that are
incorrect based strictly on the provided guideline.

You MUST format EACH of the 10 items exactly as follows, using these
specific XML tags:

<qa>
<question>
Patient scenario and the specific question here.
A) [Option 1]
B) [Option 2]
C) [Option 3]
D) [Option 4]
</question>
<answer>The rationale explaining your chain of thought without
mentioning the guideline and then Answer: correct answer</answer>
</qa>
\end{verbatim}
\end{tcolorbox}

\subsection{Synthetic Curated QA Prompt}
\label{appendix:prompt_curated}

The Synthetic Curated QA pipeline samples five exemplars without replacement from the curated benchmark pool and produces a single new QA pair per call. The pool is partitioned into labeled (multiple-choice, carrying a \texttt{label\_letter}) and unlabeled (open-ended) buckets, and the user message is specialized accordingly.

\begin{tcolorbox}[breakable, colback=gray!5, colframe=gray!50, title=Developer Message --- Synthetic Curated QA]
\begin{verbatim}
You are an expert medical educator and physician tasked with creating
high-quality, clinically accurate content for a medical exam. Your task
is to generate a new, unique, clinical vignette-style question along
with its evidence-based correct answer. The timeline and progress
should always be clear and detailed in the vignettes. Include clear
context about site and where people travelled etc. The content must
reflect realistic clinical scenarios, standard-of-care protocols, and
well established medical consensus. Avoid scientifically controversial
treatments. You will be provided with 5 examples. Use them strictly to
understand the desired format, diagnostic difficulty, and clinical
depth. DO NOT copy them. Generate a completely new, scientifically
rigorous question that would be unconditionally approved by a medical
review board.
\end{verbatim}
\end{tcolorbox}

\begin{tcolorbox}[breakable, colback=gray!5, colframe=gray!50, title=User Message --- Synthetic Curated QA]
\begin{verbatim}
Here are example questions and answers to model your format on:

--- Example 1 ---
<question>{question_1}
<answer>{answer_1}

--- Example 2 ---
<question>{question_2}
<answer>{answer_2}

... (5 examples total) ...

# If sampled from the labeled bucket:
Now generate a brand new, unique, and clinically accurate
MULTIPLE-CHOICE medical question and its detailed answer.
Ensure the answer matches the formatting tags above.

# If sampled from the unlabeled bucket:
Now generate a brand new, unique, and clinically accurate
OPEN-ENDED medical question and its detailed answer.
Ensure the answer matches the formatting tags above.
\end{verbatim}
\end{tcolorbox}

\subsection{Synthetic MOOVE Prompt}
\label{appendix:prompt_moove}

The Synthetic MOOVE pipeline samples five exemplar prompts without replacement from the MOOVE training split and generates a single new open-ended clinical scenario per call. Only the question stem is generated; assistant responses are produced downstream.

\begin{tcolorbox}[breakable, colback=gray!5, colframe=gray!50, title=Developer Message --- Synthetic MOOVE]
\begin{verbatim}
You are an expert medical educator and physician tasked with creating
high-quality, clinically accurate content. Your task is to generate a
new, unique, and realistic medical scenario or question prompt. The
content must reflect realistic clinical presentations, inquiries from
colleagues, or patient encounters. The timeline and objective progress
should always be clear and detailed. Include clear context about site
and where people travelled etc. You will be provided with 5 examples.
Use them strictly to understand the desired format, diagnostic
difficulty, and clinical depth. DO NOT copy them. Generate a completely
new question that would be unconditionally approved by a medical review
board.
\end{verbatim}
\end{tcolorbox}

\begin{tcolorbox}[breakable, colback=gray!5, colframe=gray!50, title=User Message --- Synthetic MOOVE]
\begin{verbatim}
Here are example prompts to model your format and clinical depth on:

--- Example 1 ---
<question>
{prompt_1}
</question>

--- Example 2 ---
<question>
{prompt_2}
</question>

... (5 examples total) ...

Now, acting as an expert medical educator, generate a brand new, unique,
and clinically accurate medical scenario or question. Wrap your
generated scenario strictly within <question> and </question> tags.
\end{verbatim}
\end{tcolorbox}

\section{Decontamination details}
\label{appendix:decontamination}

We apply a two-stage n-gram and token-alignment decontamination pipeline adapted from Apertus~\cite{apertus2025}\footnote{\url{https://github.com/swiss-ai/posttraining-data/tree/main/04-decontamination}} to the full curated corpus. The reference set aggregates the prompts of all evaluation benchmarks used in this work: MedQA, MedMCQA, PubMedQA, MedXpertQA, MMLU-Pro, IFEval, and ARC-Challenge.

Training samples are tokenized with \texttt{alehc/swissai-tokenizer}. In the first stage, samples sharing any 8-gram with a reference prompt are flagged as candidates. In the second stage, each candidate is token-aligned against the matched reference and removed if the normalized alignment difference is at most $\tau = 0.5$. This filters incidental n-gram overlaps while still catching lightly paraphrased contaminations. For each dataset, the pipeline outputs a decontaminated corpus and a report logging removed samples and their matched references.

\section{Full openness comparison across medical LLMs}
\label{appendix:fully_open_comparison}

We assess all models discussed in this work along four openness dimensions defined by the Model Openness Framework (MOF)~\cite{white2024model}: released weights, publicly available training data, a reproducible training recipe, and medical specialisation. As argued in Section~\ref{sec:related}, releasing weights alone does not constitute full openness: the pretraining data and training recipe of the base model determine what clinical knowledge and biases the model has absorbed, yet these dimensions are undisclosed for every major open-weight base used in prior medical LLM work (Llama 2, Llama 3.1, Mistral, Qwen2.5, Gemma 3). Apertus is the first base model at this scale to satisfy all MOF dimensions simultaneously, and Apertus-MeditronFO inherits this property while adding medical specialisation.

\begin{table}[H]
\centering
\caption{Openness dimensions across medical LLMs and their base models, following the
Model Openness Framework. \textbf{YES} = fully satisfied;
\textbf{\textasciitilde} = partial (e.g.\ some data disclosed, recipe absent);
\textbf{NO} = not satisfied. ``Medical specialist'' denotes a model adapted for clinical
tasks via continued pretraining or supervised fine-tuning on medical data.}
\label{tab:fully_open_comparison1}
\resizebox{\textwidth}{!}{%
\begin{tabular}{lllcccc}
\toprule
\textbf{Model} & \textbf{Base model} & \textbf{Base openness} &
\textbf{Open weights} & \textbf{Open data} & \textbf{Open training} &
\textbf{Medical specialist} \\
\midrule
\multicolumn{7}{l}{\textit{Closed specialists}} \\
\addlinespace[2pt]
MedPaLM & PaLM (closed) & Closed & NO & NO & NO & YES \\
MedPaLM-2 & PaLM 2 (closed) & Closed & NO & NO & NO & YES \\
Med-Gemini & Gemini (closed) & Closed & NO & NO & NO & YES \\

\addlinespace[2pt]
\midrule
\addlinespace[2pt]
\multicolumn{7}{l}{\textit{Open-weight medical specialists}} \\
\addlinespace[2pt]
PMC-LLaMA & Llama 2 (open-wt) & Open-weight & YES & NO & NO & YES \\
BioMistral & Mistral (open-wt) & Open-weight & YES & NO & NO & YES \\
Meditron-70B & Llama 2 (open-wt) & Open-weight & YES & NO & NO & YES \\
HuatuoGPT-o1 & Qwen2.5 (open-wt) & Open-weight & YES & NO & NO & YES \\
ReasonMed & Qwen2.5 (open-wt) & Open-weight & YES & YES & NO & YES \\
Aloe Beta & Llama 3.1 / Qwen2.5 & Open-weight & YES & YES & NO & YES \\
Hippocrates & Llama 2 / Mistral & Open-weight & YES & YES & NO & YES \\
MediPhi & Phi-3.5 (open-wt) & Open-weight & YES & NO & NO & YES \\
Llama-3-Meditron & Llama 3.1 (open-wt) & Open-weight & YES & NO & NO & YES \\
MedGemma-27B & Gemma 3 (open-wt) & Open-weight & YES & NO & NO & YES \\
\addlinespace[2pt]
\midrule
\addlinespace[2pt]
\multicolumn{7}{l}{\textit{Fully open specialists (this work)}} \\
\addlinespace[2pt]
Apertus-70B-MeditronFO  & Apertus-70B & \textbf{Fully open} & \textbf{YES} & \textbf{YES} & \textbf{YES} & \textbf{YES} \\
Apertus-8B-MeditronFO   & Apertus-8B  & \textbf{Fully open} & \textbf{YES} & \textbf{YES} & \textbf{YES} & \textbf{YES} \\
OLMo-2-32B-MeditronFO   & OLMo-2-32B   & \textbf{Fully open} & \textbf{YES} & \textbf{YES} & \textbf{YES} & \textbf{YES} \\
EuroLLM-22B-MeditronFO  & EuroLLM-22B  & \textbf{Fully open} & \textbf{YES} & \textbf{YES} & \textbf{YES} & \textbf{YES} \\
EuroLLM-9B-MeditronFO   & EuroLLM-9B   & \textbf{Fully open} & \textbf{YES} & \textbf{YES} & \textbf{YES} & \textbf{YES} \\
\bottomrule
\end{tabular}}
\end{table}

\begin{figure}[h]
    \centering
    \includegraphics[width=\linewidth]{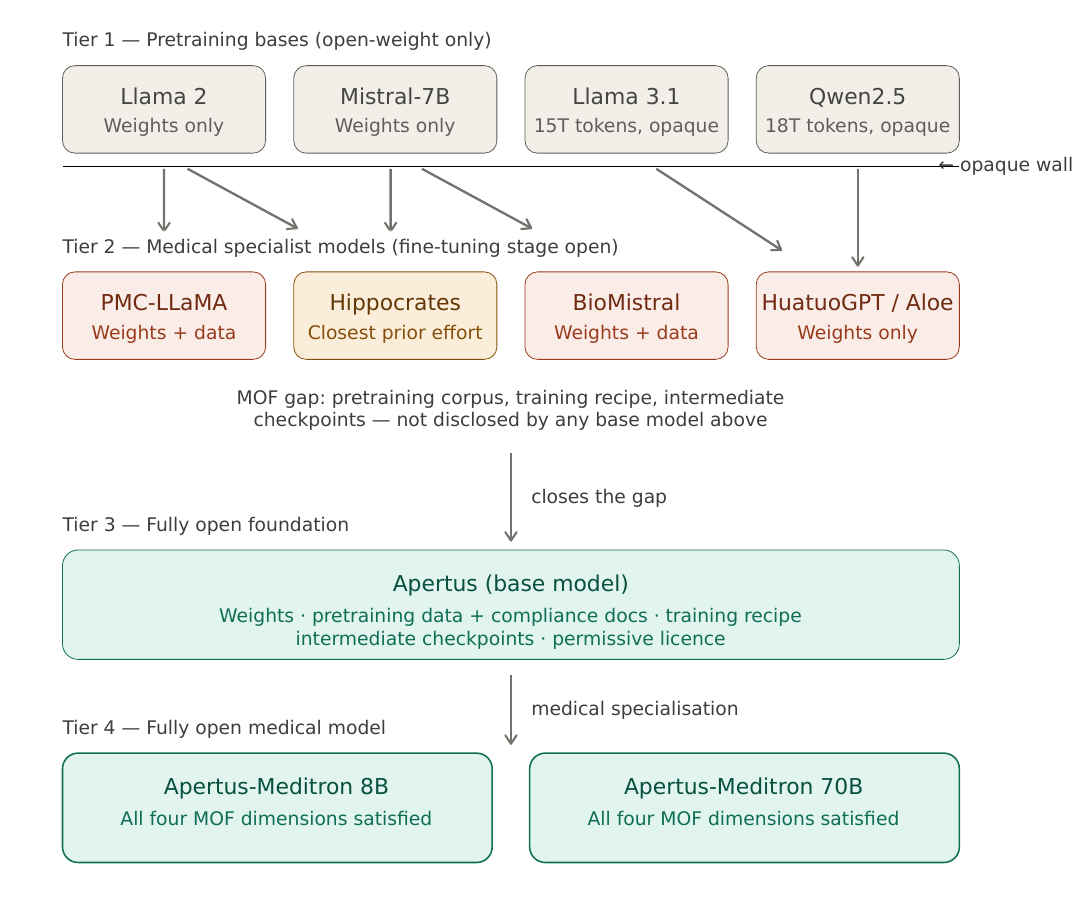}
    \caption{Medical LLM Openness Tiers}
    \label{fig:myplot}
\end{figure}

\section{Licenses of existing assets}
\label{app:licenses}

Table~\ref{tab:asset_licenses} lists each existing asset used in this work, with its originating reference, version where applicable, public URL, and license. All assets are used in accordance with their respective licenses and restricted to research purposes consistent with the originating works.

\begin{table}[H]
\centering
\caption{Licenses of existing assets used in this work. Datasets and models are listed with their originating reference, version, public URL, and license. †~License terms verified at time of writing; downstream users should re-verify upstream terms before redistribution.}
\label{tab:asset_licenses}
\resizebox{\textwidth}{!}{%
\begin{tabular}{llll}
\toprule
\textbf{Asset} & \textbf{Reference} & \textbf{URL} & \textbf{License} \\
\midrule
\multicolumn{4}{l}{\textit{Source QA datasets (training)}} \\
\midrule
MedQA              & Jin et al. [2021]      & \url{https://github.com/jind11/MedQA}                  & MIT \\
MedMCQA            & Pal et al. [2022]      & \url{https://medmcqa.github.io}                        & MIT \\
PubMedQA           & Jin et al. [2019]      & \url{https://pubmedqa.github.io}                       & MIT \\
MedExpQA           & Alonso et al. [2024]   & \url{https://huggingface.co/datasets/HiTZ/MedExpQA}    & CC BY-NC-SA 4.0 \\
HealthSearchQA     & Singhal et al. [2023]  & via Med-PaLM release                                   & CC BY 4.0 \\
LiveQA-Med         & Abacha et al. [2017]   & \url{https://github.com/abachaa/LiveQA_MedicalTask_TREC2017} & Open / research use\textsuperscript{†} \\
AfriMed-QA v1/v2   & Olatunji et al. [2024] & \url{https://huggingface.co/datasets/intronhealth/afrimedqa_v2} & CC BY 4.0\textsuperscript{†} \\
GUIDELINES corpus  & Chen et al. [2023b]    & via Meditron release                                   & Per-source (mixed); research use\textsuperscript{†} \\
MOOVE (training split) & Sallinen et al. [2025] & via Llama-3-Meditron release                       & Research use\textsuperscript{†} \\
\midrule
\multicolumn{4}{l}{\textit{Evaluation benchmarks}} \\
\midrule
MedXpertQA         & Zuo et al. [2025]      & \url{https://huggingface.co/datasets/TsinghuaC3I/MedXpertQA} & MIT\textsuperscript{†} \\
MMLU-Pro           & Wang et al. [2024]     & \url{https://huggingface.co/datasets/TIGER-Lab/MMLU-Pro}    & MIT \\
IFEval             & Zhou et al. [2023]     & \url{https://github.com/google-research/google-research/tree/master/instruction_following_eval} & Apache 2.0 \\
ARC-Challenge      & Clark et al. [2018]    & \url{https://allenai.org/data/arc}                     & CC BY-SA 4.0 \\
HealthBench        & Arora et al. [2025]    & \url{https://github.com/openai/simple-evals}           & MIT\textsuperscript{†} \\
\midrule
\multicolumn{4}{l}{\textit{Base models (fine-tuned)}} \\
\midrule
Apertus-70B-Instruct  & Hernández-Cano et al. [2025] & \url{https://huggingface.co/swiss-ai/Apertus-70B-Instruct-2509} & Apache 2.0 \\
Apertus-8B-Instruct   & Hernández-Cano et al. [2025] & \url{https://huggingface.co/swiss-ai/Apertus-8B-Instruct-2509}  & Apache 2.0 \\
OLMo-2-32B-SFT        & OLMo et al. [2024]           & \url{https://huggingface.co/allenai/OLMo-2-0325-32B-SFT}        & Apache 2.0 \\
EuroLLM-22B-Instruct  & Ramos et al. [2026]          & \url{https://huggingface.co/utter-project/EuroLLM-22B-Instruct} & Apache 2.0\textsuperscript{†} \\
EuroLLM-9B-Instruct   & Ramos et al. [2026]          & \url{https://huggingface.co/utter-project/EuroLLM-9B-Instruct}  & Apache 2.0 \\
Gemma-3-27B-IT        & Team et al. [2025]           & \url{https://huggingface.co/google/gemma-3-27b-it}              & Gemma Terms of Use \\
\midrule
\multicolumn{4}{l}{\textit{Reference / baseline models (evaluation only)}} \\
\midrule
MedGemma-27B           & Sellergren et al. [2025] & \url{https://huggingface.co/google/medgemma-27b-text-it} & Health AI Developer Foundations TOS \\
Llama-3.1-70B-Meditron & Sallinen et al. [2025]   & \url{https://huggingface.co/OpenMeditron/Meditron3-70B}  & Llama 3.1 Community License \\
MediPhi                & Corbeil et al. [2025]    & \url{https://huggingface.co/microsoft/MediPhi}           & MIT\textsuperscript{†} \\
Qwen3-30B-A3B-Instruct-2507 & Yang et al. [2025]  & \url{https://huggingface.co/Qwen/Qwen3-30B-A3B-Instruct-2507} & Apache 2.0 \\
Qwen3-235B-A22B        & Yang et al. [2025]       & \url{https://huggingface.co/Qwen/Qwen3-235B-A22B}        & Apache 2.0 \\
gpt-oss-120b           & Agarwal et al. [2025]    & \url{https://huggingface.co/openai/gpt-oss-120b}         & Apache 2.0 \\
\midrule
\multicolumn{4}{l}{\textit{Frameworks and infrastructure}} \\
\midrule
Axolotl                & --- & \url{https://github.com/axolotl-ai-cloud/axolotl} & Apache 2.0 \\
PyTorch (FSDP v2)      & --- & \url{https://pytorch.org}                         & BSD 3-Clause \\
DeepSpeed (ZeRO-3)     & --- & \url{https://github.com/deepspeedai/DeepSpeed}    & Apache 2.0 \\
FlashAttention 2       & --- & \url{https://github.com/Dao-AILab/flash-attention}& BSD 3-Clause \\
\bottomrule
\end{tabular}}
\end{table}

%%%%%%%%%%%%%%%%%%%%%%%%%%%%%%%%%%%%%%%%%%%%%%%%%%%%%%%%%%%%

%\newpage
%\input{checklist.tex}

\end{document}